\documentclass[10pt,twocolumn,letterpaper]{article}

\usepackage[pagenumbers]{cvpr} 
\usepackage{subfiles}
\usepackage{savesym}

\usepackage{amsmath}
\usepackage{amssymb}
\usepackage[linesnumbered,ruled,vlined]{algorithm2e}
\usepackage{tabularx,booktabs}
\SetKwInOut{KwInput}{Input}
\SetKwInOut{KwOutput}{Output}
\usepackage{color}
\usepackage{colortbl}
\usepackage{multirow}
\usepackage{multicol}
\usepackage{subcaption}
\usepackage{graphicx}
\usepackage{lipsum}  
\usepackage{makecell}
\usepackage{xcolor}
\usepackage{subcaption}
\usepackage{graphicx}
\SetKwInput{KwRequire}{Require}
\usepackage[pagebackref,breaklinks,colorlinks]{hyperref}

\usepackage[capitalize]{cleveref}
\crefname{section}{Sec.}{Secs.}
\Crefname{section}{Section}{Sections}
\Crefname{table}{Table}{Tables}
\crefname{table}{Tab.}{Tabs.}

 \def\hlinewd#1{%
      \noalign{\ifnum0=`}\fi\hrule \@height #1 \futurelet
      \reserved@a\@xhline}

\newcommand{\figref}[1]{Fig. \ref{#1}}
\newcommand{\tabref}[1]{Table \ref{#1}}
\newcommand{\equref}[1]{Eq. (\ref{#1})}
\newcommand{\secref}[1]{Sec. \ref{#1}}
\newcommand\blfootnote[1]{%
  \begingroup
  \renewcommand\thefootnote{}\footnote{#1}%
  \addtocounter{footnote}{-1}%
  \endgroup
}      
\def\cvprPaperID{4868} 
\def\confName{CVPR}
\def\confYear{2023}
\begin{document}

\documentclass[10pt,twocolumn,letterpaper]{article}

\usepackage[pagenumbers]{cvpr} 
\usepackage{savesym}
\usepackage{epstopdf}


\usepackage{amsmath}
\usepackage{amssymb}
\usepackage[linesnumbered,ruled,vlined]{algorithm2e}
\usepackage{tabularx,booktabs}
\SetKwInOut{KwInput}{Input}
\SetKwInOut{KwOutput}{Output}
\usepackage{color}
\usepackage{colortbl}
\usepackage{multirow}
\usepackage{multicol}
\usepackage{subcaption}
\usepackage{graphicx}
\usepackage{lipsum}  
\usepackage{makecell}
\usepackage{xcolor}
\usepackage{subcaption}
\usepackage{graphicx}
%
\usepackage[pagebackref,breaklinks,colorlinks]{hyperref}
\epstopdfsetup{outdir=./}

\usepackage[capitalize]{cleveref}
\crefname{section}{Sec.}{Secs.}
\Crefname{section}{Section}{Sections}
\Crefname{table}{Table}{Tables}
\crefname{table}{Tab.}{Tabs.}

 \def\hlinewd#1{%
      \noalign{\ifnum0=`}\fi\hrule \@height #1 \futurelet
      \reserved@a\@xhline}

\newcommand{\figref}[1]{Fig. \ref{#1}}
\newcommand{\tabref}[1]{Table \ref{#1}}
\newcommand{\equref}[1]{Eq. (\ref{#1})}
\newcommand{\secref}[1]{Sec. \ref{#1}}
\newcommand\blfootnote[1]{%
  \begingroup
  \renewcommand\thefootnote{}\footnote{#1}%
  \addtocounter{footnote}{-1}%
  \endgroup
}      
\def\cvprPaperID{4868} 
\def\confName{CVPR}
\def\confYear{2023}
\def\supplementfilename{supplement.pdf}
\pdfximage{\supplementfilename}
\def\numbersupplementpages{\the\pdflastximagepages}
\newif\ifarXiv
\arXivtrue 
\begin{document}

\title{Probabilistic Prompt Learning for Dense Prediction}
\vspace{-20pt}
\author{Hyeongjun Kwon$^1$ \quad Taeyong Song$^2$\quad Somi Jeong$^3$ \quad Jin Kim$^1$ \quad  Jinhyun Jang$^1$ \quad Kwanghoon Sohn$^{1, 4}$\thanks{Corresponding author.}\\
$^1$Yonsei University, \quad $^2$Hyundai Motor Company R\&D Division,\\
$^3$NAVER LABS, \quad  $^4$Korea Institute of Science and Technology (KIST)\\
{\tt\small \{kwonjunn01, kimjin928, jr000192, khsohn\}@yonsei.ac.kr},\\ {\tt\small taeyongsong@hyundai.com, somi.jeong@naverlabs.com}}
\maketitle
\blfootnote{This research was supported by the National Research Foundation of Korea (NRF) grant funded by the Korea government (MSIP) (NRF2021R1A2C2006703).}
\begin{abstract}
Recent progress in deterministic prompt learning has become a promising alternative to various downstream vision tasks, enabling models to learn powerful visual representations with the help of pre-trained vision-language models.
However, this approach results in limited performance for dense prediction tasks that require handling more complex and diverse objects, since a single and deterministic description cannot sufficiently represent the entire image.
In this paper, we present a novel probabilistic prompt learning to fully exploit the vision-language knowledge in dense prediction tasks. 
First, we introduce learnable class-agnostic attribute prompts to describe universal attributes across the object class. 
The attributes are combined with class information and visual-context knowledge to define the class-specific textual distribution. 
Text representations are sampled and used to guide the dense prediction task using the probabilistic pixel-text matching loss, enhancing the stability and generalization capability of the proposed method.
Extensive experiments on different dense prediction tasks and ablation studies demonstrate the effectiveness of our proposed method.
\end{abstract}\vspace{-15pt}

\section{Introduction}
Dense predictions, \eg, semantic segmentation~\cite{chen2017deeplab,long2015fully}, instance segmentation~\cite{lin2014microsoft}, and object detection~\cite{girshick2014rich,ren2015faster}, are fundamental computer vision problems, which aim to produce pixel-level predictions to thoroughly understand the scene.
Due to the expensive cost of collecting dense annotations, most approaches~\cite{ghiasi2018dropblock,poudel2019fast} employ a \textit{``pre-training + fine-tuning''} paradigm.
Based on existing pre-trained networks~\cite{he2016deep,dosovitskiy2020vit} trained on large-scale datasets such as ImageNet~\cite{deng2009imagenet}, semi-~\cite{ouali2020semi,zhou2020learning} or self-supervised learning~\cite{zhan2018mix,li2020improving} has been extensively researched to fine-tune additional modules for dense prediction.
However, due to the biased visual representations, they suffer from a lack of scalability when there exists a large semantic gap between pre-trained and target tasks \wrt dataset and objective, such as transferring ImageNet classification network to COCO object detection~\cite{he2019rethinking,zoph2020rethinking}.

\begin{figure}[t]
    \centering
    \includegraphics[width=0.95\linewidth]{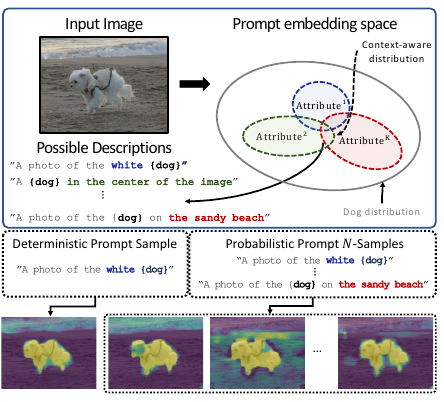}
    \vspace{-5pt}
    \caption{\textbf{Probabilistic Prompt Learning.}
The proposed PPL exploits $N$ multiple prompts sampled from probabilistic text embeddings, which can leverage granular textual representations, enabling a better understanding of the details of the image.
} 
\vspace{-15pt}
\label{fig:concept}
\end{figure}

Recently, applying vision-language pre-trained models (VLM) to the downstream tasks has demonstrated remarkable success, including zero-shot classification~\cite{zhou2022conditional,zhou2022learning}, referring expression segmentation~\cite{wang2022cris}, and object detection~\cite{rao2022denseclip}.
VLM, such as CLIP~\cite{radford2021learning} and ALIGN~\cite{jia2021scaling}, is trained on large-scale web noisy image-text pair datasets via contrastive learning to align representations between text and image in a joint embedding space.
In this way, VLM learns robust visual representations by exploiting the semantic relationship between text and image representations, which is beneficial to transfer knowledge to various vision tasks.
To efficiently leverage the language knowledge, there have been many pioneering attempts~\cite{wang2022dualprompt,ju2021prompting,he2022towards} to deploy VLM to the downstream tasks via \emph{prompting}.
For example, based on a prompt template \texttt{"a photo of a \{class\}"}, it measures the confidence score by calculating image-text similarity and classifies the image into \texttt{\{class\}}.
In practice, however, this hand-crafted prompt may not be the optimal description for a particular task, and furthermore, manually designing a task-specific prompt is laborious.

To tackle this issue, several methods~\cite{zhou2022learning,zhou2022conditional,rao2022denseclip} have introduced learning-based prompting techniques inspired by early works in NLP~\cite{houlsby2019parameter,li2021prefix}.
The goal is to automatically construct the prompts according to the task by optimizing continuous prompt variables based on VLM.
This simple and intuitive setup, referred to as \textit{deterministic prompt learning}, is the most popular approach in the current literature~\cite{feng2022promptdet,wang2022dualprompt}.
It has shown performance improvement on classification tasks~\cite{radford2021learning,Zhai_2022_CVPR}, where a single deterministic embedding is sufficient for representing a class.
However, this approach is not fully compositional in dense prediction tasks due to a semantic ambiguity problem.
Firstly, while the dense prediction tasks require granular information to generate precise pixel-wise results, not only complex and multiple objects within an input image but also their various attributes (\eg, color, location, etc.) cannot be comprehensively represented in a single textual representation~\cite{hao2019collect,yang2020prototype}.
Thus, a single prompt fails to comprehend the object of all classes in detail.
Second, the visual representation has high randomness~\cite{chun2021pcme, ding2021vision} due to various contexts with external objects and object representations, and it results in high uncertainty in representing in language.
For example, as shown in Fig.~\ref{fig:concept}, the image can be described as \texttt{"A photo of the dog on the sandy beach"}, but it can also be expressed differently such as \texttt{"A photo of the dog near the ocean"}.
Therefore, it is not appropriate to exploit a deterministically visual representation when transferring VLM in dense prediction tasks.

In this paper, we propose a Probabilistic Prompt Learning (\textbf{PPL}) that explores learning appropriate textual descriptions using visual cues in a probabilistic embedding space.
We present a set of prompts that express class-agnostic attributes such as position, size, and color to represent objects without semantic ambiguity.
To effectively learn the probabilistic distribution to describe diverse and informative attributes for the entire class, we model each attribute distribution as a distinct normal distribution.
To this end, we set its variance as contextual relations between text and visual features to explicitly utilize visual-text information.
With these attribute distributions, class-specific attribute distribution is approximated by a Mixture of Gaussian (MoG).
Furthermore, we introduce a novel probabilistic pixel-text matching loss to attenuate the instability of prediction probability caused by high uncertainty.

In summary, our contributions are as follows: (1) We propose a novel PPL to effectively represent class-agnostic attributes of objects in probabilistic text embedding space.
To the best of our knowledge, this is the first attempt to leverage context-aware probabilistic prompt learning.
(2) We introduce a novel probabilistic pixel-text matching loss to alleviate the adverse impact of uncertainty.
(3) We demonstrate the effectiveness of the proposed probabilistic approach through extensive experiments on dense prediction tasks, including semantic segmentation, instance segmentation, and object detection.
\vspace{-8pt}
\section{Related Works}
\vspace{-5pt}
\paragraph{Vision-Language pre-trained Model.}
Vision-language pre-trained models (VLM)~\cite{hong2021vln,huang2021seeing,kamath2021mdetr,kim2021vilt} have been widely researched on various downstream vision tasks,  including visual question answering~\cite{alberti2019fusion}, image captioning~\cite{xia2021xgpt}, text-to-image retrieval~\cite{qi2020imagebert}, and so on.
Conventional VLM learns the connections between image content and language via extra modules.
However, due to the relatively small dataset, most approaches had difficulty aligning the connection between image and language.
Recently, applying contrastive learning on noisy web-scale image-text paired datasets has shown promising results in VLM such as CLIP~\cite{radford2021learning} and ALIGN~\cite{jia2021scaling}.
Leveraging this language supervision from VLM, there exists remarkable improvement in various downstream vision tasks under unannotated or restricted data regimes.
\vspace{-10pt}
\paragraph{Prompt Learning.}
Prompt learning, inspired by the concurrent work in NLP~\cite{li2021prefix}, is widely researched for VLM, which aims to learn to generate optimal descriptions that enhance the visual-text representations.
CoOp~\cite{zhou2022learning} is a pioneer work that applied prompt learning to vision tasks, and it leverages learnable continuous prompts trained on the freeze CLIP encoder.
Lu \etal~\cite{lu2022prompt} proposed ProDA to optimize multiple sets of prompts by learning the distribution of prompts, which discovers task-related content with less bias than manual design.
With the recent progress of prompt learning, these approaches have demonstrated impressive improvements in high-level vision tasks, including video recognition~\cite{ni2022expanding,ju2021prompting}, point cloud understanding~\cite{zhang2022pointclip}, and dense prediction~\cite{rao2022denseclip}.
Especially, DenseCLIP~\cite{rao2022denseclip} is proposed to apply that prompt learning to dense prediction tasks, where pixel-text matching loss is used as a guide for the task loss.
Despite the progress of prompting leveraging the visual-context, they still do not consider the randomness of the visual-context.
\vspace{-10pt}
\paragraph{Probabilistic Embedding.}
Learning probabilistic representations is a traditional approach in the word embedding approach~\cite{wang2021enhancing}.
Since they fully exploit the inherent hierarchical structure in language embeddings, it has been widely studied for structuring different distributions with word representations.
Recently, probabilistic embedding approaches have received more attention in the field of computer vision.
Hedged Instance Embedding (HIB)~\cite{oh2018modeling} is proposed to handle one-to-many mappings via the variational information bottleneck (VIB) principle.
Based on the HIB concept, probabilistic embeddings for cross-modal retrieval~\cite{chun2021pcme} have been studied to learn joint embeddings to capture one-to-many associations.
\vspace{-10pt}
\paragraph{Uncertainty in Computer Vision.}
Uncertainty is one of the main problems that degrade reliable performance in CNN-based methods.
Therefore, various methods of handling this uncertainty to improve robustness and performance have been extensively studied in various applications such as image retrieval~\cite{chun2021pcme}, face recognition~\cite{chang2020data}, video retrieval~\cite{park2022probabilistic}, and semantic segmentation~\cite{kendall2017uncertainties}.
In general, uncertainty can be classified into two types depending on its cause:
(1) Epistemic uncertainty, called model uncertainty, is derived from model parameters. 
(2) Aleatoric uncertainty, called data uncertainty, is originated from the inherent noise of data.
Epistemic uncertainty can be reduced by providing sufficient data, while aleatoric uncertainty is irreducible with supplementary data.
Although some works~\cite{vilnis2014word,yu2019robust,yang2021probabilistic,chun2021pcme,park2022probabilistic} have attempted to define uncertainty through the variance from their dataset and handle it, there is no direct applicable approach to VLM in conventional computer vision tasks due to the lack of linguistic datasets.
In this work, we explore the aleatoric uncertainty of language with only the image-modality dataset. 
\label{sec:relatedwork} 
\vspace{-5pt}
\section{Background}
\vspace{-5pt}
\paragraph{Contrastive Language-Image Pre-training (CLIP)}\cite{radford2021learning} is a powerful vision-language pre-trained model that learns to align image-text representations via contrastive learning~\cite{oord2018representation}.
It considers the relevant image and class text description pairs $\{\mathbf{x}, \mathbf{t}_c\}$ as positive samples and the non-relevant pairs as negative samples.
The contrastive objective is to maximize the similarity of the positive samples while minimizing the similarity of negative samples.
To be specific, it consists of an image encoder $\mathcal{F}$ and a text encoder $\mathcal{G}$.
Given $\mathbf{x}$, it measures the similarity between image and class-embedded text representations $\boldsymbol{w}_{1:C}=\{w_1,...,w_C\}$, where $w_c=\mathcal{G}(\mathbf{t}_c)$ and $C$ denotes the number of classes.
Then, the prediction probability that $\mathbf{x}$ belongs to class $c$ is computed as follows:
\begin{equation}
   p(y_c|\mathbf{x}, \boldsymbol{w}_{1:C}) = \frac{\exp({\mathcal{F}{(\mathbf{x})\cdot {w}_{c}^\top/\tau}})}{\sum_{i=1}^{C}\exp(\mathcal{F}{(\mathbf{x})\cdot {w}_{i}^\top/\tau})}, \\
\end{equation}
where $\tau$ is a hyper-parameter.

\vspace{-10pt}
\paragraph{Context Optimization (CoOp)}\cite{zhou2022learning} introduced a learning-based prompt method that learns task-relevant prompts for better transferability in downstream recognition tasks.
The prompt $\mathbf{p}$ is revised as learnable continuous variables, which is updated by back-propagation with the pre-trained CLIP text encoder.
In particular, the class description with $\mathbf{p}$ is obtained by concatenating their text token as $\mathbf{t}_c(\mathbf{p})$.
It can optimize $\mathbf{p}$ using $M$ training samples $\{\mathbf{x}_{i},y_{i}\}_{i=1}^M$, by minimizing the following objective:
\begin{equation}
    \mathcal{L}(\mathbf{p}) = \underset{\mathbf{x}_{i},y_{i}}{\mathbb{E}}\left[-\log \,p(y_{i}|\mathbf{x}_{i}, \boldsymbol{w}_{1:C}(\mathbf{p})\right],
\end{equation}
where $\boldsymbol{w}_{1:C}(\mathbf{p})=\{w_1(\mathbf{p}), ..., w_C(\mathbf{p})\}$ and $w_c(\mathbf{p})=\mathcal{G}(\mathbf{t}_c(\mathbf{p}))$.

\vspace{-10pt}
\paragraph{Prompt Distribution Learning (ProDA)}\cite{lu2022prompt} aims to learn the distribution of diverse prompts to handle various visual representations.
They assume that the prompt distribution $p(\mathbf{p})$ can be modeled as a Gaussian distribution.
Specifically, they define a set of $K$ learnable prompts as $\mathbf{P}=\{\mathbf{p}^1,...,\mathbf{p}^K\}$, and model the distribution of $\boldsymbol{w}_{1:C}(\mathbf{P}) \sim \mathcal{N}(\boldsymbol{\mu}_{{w}_{1:C}(\mathbf{P})}, \boldsymbol{\sigma}_{{w}_{1:C}(\mathbf{P})}\mathbf{I})$ with a mean and diagonal covariance of $\boldsymbol{w}_{1:C}(\mathbf{P})=\{\boldsymbol{w}_{1:C}(\mathbf{p}^1), ...,$ $\boldsymbol{w}_{1:C}(\mathbf{p}^K)\}$.
To learn an optimal prompt distribution, $\mathbf{P}$ is optimized by minimizing the classification loss as:
\begin{equation}
    \mathcal{L}(\mathbf{P}) = \underset{\mathbf{x}_{i},y_{i}}{\mathbb{E}}\left[-\log \underset{\boldsymbol{w}_{1:C}(\mathbf{P})}{\mathbb{E}}\left[p(y_{i}|\mathbf{x}_{i}, \boldsymbol{w}_{1:C}(\mathbf{P})\right]\right].
\end{equation}
While the ProDA has achieved state-of-the-art results in a large variety of downstream tasks, it is vulnerable to generating context-aware text representations.

\vspace{-8pt}
\paragraph{DenseCLIP} \cite{rao2022denseclip} leverages the pixel-text matching loss $\mathcal{L}_{pixel}$ as a guidance of task objective. 
In particular, DenseCLIP suggests a post-model prompting approach, which directly adds visual-context bias $u = \mathrm{TransDecoder}(\boldsymbol{w}_{1:C}(\mathbf{p}), \mathcal{F}(\mathbf{x}))$ to $\boldsymbol{w}_{1:C}$.
It obtains the context-aware text representations as $\hat{\boldsymbol{w}}_{1:C} \xleftarrow{} \boldsymbol{w}_{1:C} + \gamma u$,
where $\gamma$ is a learnable scale factor.
Then, they utilize pixel-text contrastive loss as an auxiliary loss for task-specific loss, which is formulated follow as:
\begin{equation}
    \mathcal{L}_{pixel} = \underset{\mathbf{x}_{i,j},y_{c}}{\mathbb{E}}\left[-\log p(y_{c}|\mathbf{x}_{i,j}, \hat{\boldsymbol{w}}_{1:C})\right],
\end{equation}
where $i,j$ are each pixel location of input image $\mathbf{x}$.
Although they utilize visual cues as context bias via context-aware prompting, a deterministic prompt is still not enough to deal with the randomness of visual representations.

\begin{figure*}[t]
		\centering
		\includegraphics[width=0.99\linewidth]{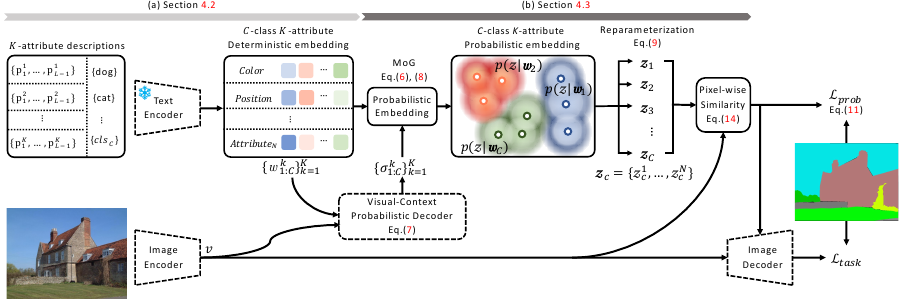}
  \vspace{-10pt}
	    \caption{\textbf{Overall framework of PPL}. (a) The text encoder takes $K$-attribute prompts to generate the deterministic text embeddings $\boldsymbol{w}_c$ for each class $c$. The standard deviation $\sigma_c^k$ of class-attribute embedding $w_c^k$ is then computed via the visual-context probabilistic decoder. (b) Each class distribution is modeled as a Mixture of Gaussian (MoG), from which the text embeddings $\boldsymbol{z}_c$ are randomly sampled.}\vspace{-15pt}
         \label{fig:framework}
\end{figure*}
\vspace{-5pt}
\section{Method}
\vspace{-3pt}
\subsection{Overview} \label{sec:overview}
\vspace{-3pt}
In this section, we present our proposed Probabilistic Prompt Learning (PPL) for dense predictions, illustrated in Fig. \ref{fig:framework}. 
Given an image $\mathbf{x}$, our goal is to predict plausible pixel-wise results $y$ by taking advantage of general knowledge learned from VLM. 
It consists of an image encoder $\mathcal{F}$, a text encoder $\mathcal{G}$, and an image decoder $\mathcal{D}$ for generating the dense prediction results.
We utilize multiple text representations, which are combinations of class-agnostic attribute prompts and object classes (\figref{fig:framework} (a)). 
The visual and text features are fed into the visual-context probabilistic decoder to define a probabilistic embedding space (\figref{fig:framework} (b)).
The attributes' distribution for each object class is represented as a Mixture of Gaussian (MoG), from which we sample text representations to boost the dense prediction task through pixel-text similarity loss.
\vspace{-5pt}
\subsection{Class-Agnostic Attribute Prompt}
\label{section:caap}
We first introduce class-agnostic attribute prompts to understand objects with diverse perspectives.
It aims not only to learn diverse prompts from data but also to automatically assign efficient attributes that are universally available in all object classes.
Specifically, it consists of a set of $K$ learnable prompt templates $\mathbf{P}=\{\mathbf{p}^{1}, ..., \mathbf{p}^{K}\}$, to describe various attributes across the object class.
We define an association of prompt $\mathbf{p}^k$ and class $\mathbf{t}_c$ as $\mathbf{t}_c(\mathbf{p}^k)$, and set $k$-th attribute representations as $\boldsymbol{w}_{1:C}^k=\mathcal{G}(\mathbf{t}_{1:C}(\mathbf{p}^k))$.

We further propose diversity loss to regularize each class-text representation to become different from others and prevent the multiple learned text representations from being identical.
The diversity loss $\mathcal{L}_{div}$ is formulated as:
\begin{equation}
    \mathcal{L}_{div} = \frac{1}{C}\sum_{c=1}^{C}\max(\parallel \boldsymbol{w}_{c}\boldsymbol{w}_{c}^\top - \mathbf{I} \parallel_{F}^2 - \ b, 0),
    \label{eq:diversity}
\end{equation}
where $\boldsymbol{w}_{c}=\{w_{c}^{1},...,w_{c}^{K}\}$ is the set of attribute representations of class $c$ and $\parallel \! \cdot \!\parallel_F$ denotes Frobenius norm of a matrix.
Note that we control the extent of overlap between attribute representations with a learnable vector $b\in\mathbb{R}^C$.
When $b$ is close to 0, each attribute representation is trained to be orthogonal.
The representation set $\boldsymbol{w}_c$ is used to define a distribution, from which the probabilistic prompts for the class $c$ are sampled.


\vspace{-5pt}
\subsection{Probabilistic Prompt Learning}
\label{section:PPL}
We propose probabilistic prompt learning to infer the distribution of the class-attribute representations $w_{c}^k$ that describe various visual-contexts for the target objects.
From this perspective, for each $w_{c}^k$, we define a probability distribution $p(z|w_c^k)$ as a factorized Gaussian with its center vector $\boldsymbol{\mu}_c^k$ and diagonal covariance matrix $\boldsymbol{\sigma}_c^k$:
\begin{equation}
    p(z|w_{c}^k)\sim \mathcal{N}(\boldsymbol{\mu}_c^k, \boldsymbol{\sigma}_c^k \mathbf{I}).
\end{equation}
We set the center vector $\boldsymbol{\mu}_c^k = w_{c}^k$, then exploit visual-context knowledge to calculate the covariance matrix $\boldsymbol{\sigma}_c^k$. 

\begin{figure}[t]
		\centering		\includegraphics[width=0.88\linewidth]{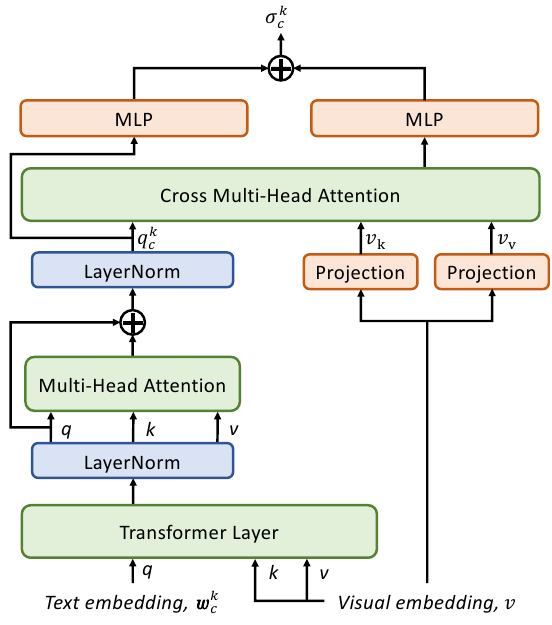}\vspace{-2pt}
	    \caption{\textbf{Architecture of the visual-context probabilistic decoder}. It takes text embedding $w_c^k$ and visual embedding $v$ as inputs, and generates a standard deviation $\sigma_c^k$ of attribute distributions $p(z|w_{c}^k)\sim \mathcal{N}(\boldsymbol{\mu}_c^k, \boldsymbol{\sigma}_c^k \mathbf{I})$.}
            \vspace{-15pt}
         \label{fig:vcprobnet}
\end{figure}

Specifically, we design a visual-context probabilistic decoder, illustrated in Fig. \ref{fig:vcprobnet}.
We feed the class-attribute representation $w_c^k$ and visual embedding $v=\mathcal{F}(\mathbf{x})$ into a transformer module~\cite{dosovitskiy2020vit}, whose output is fed into the multi-head self-attention layer to generate a new query $q_c^k$.
Here, the key and value vectors are both obtained from the visual embedding $v$.
The covariance matrix of each class-attribute representation is computed as:
\begin{equation}
    \boldsymbol{\sigma}_c^k = \mathrm{MLP}(\mathrm{LN}(q_c^k)) + \mathrm{MLP}(\mathrm{MHA}(q_c^k, v_{\mathrm{k}},v_{\mathrm{v}})),
\end{equation}
where LN and MHA are layer norm and multi-head attention.
Finally, we formulate the prompt distributions for the class as a Mixture of Gaussian (MoG) model such that
\begin{equation}\label{eq:MoG}
p(z|\boldsymbol{w}_{c})\sim \sum^K_{k=1}\mathcal{N}(\boldsymbol{\mu}^k_c,\boldsymbol{\sigma}_c^k \mathbf{I}).
\end{equation}
It can be interpreted as a distribution of possible class-attribute representations which reflects visual-context knowledge from the input image.

From $p(z|\boldsymbol{w}_c)$, we randomly sample $N$ prompt representations for class $c$ as $\boldsymbol{z}_{c}=\{z_{c}^{1},...,z_{c}^{N}\}\overset{\iid}{\sim}$ $\!p(z|\boldsymbol{w}_c)$, and we treat $\boldsymbol{z}_{c}$ as ``self-augmented'' context descriptions.
This sampling process is based on reparameterization trick~\cite{kingma2015variational} 
\begin{equation}
    z^n_{c} = \mu(\boldsymbol{w}_{c}) + \epsilon^n\sigma(\boldsymbol{w}_{c}),
\end{equation}
where $\mu(\boldsymbol{w}_{c})$, $\sigma(\boldsymbol{w}_{c})$ are mean and standard deviation of $p(z|\boldsymbol{w}_{c})$, and $\epsilon^{n} \sim \mathcal{N}(0,I)$.
Finally, we transfer the text knowledge to the dense prediction task by pixel-text matching loss $\mathcal{L}_{pixel}$:
\begin{equation}
    \mathcal{L}_{pixel} = \underset{\mathbf{x}_{i,j}, y_{c}}{\mathbb{E}}\left[-\log\underset{\boldsymbol{w}_{1:C}}{\mathbb{E}}\left[\,p(y_{c}|\mathbf{x}_{i,j}, \boldsymbol{z}_{1:C})\,\right]\right],
    \label{eq:pixel}
\end{equation}
where $i$ and $j$ denote the pixel location in the image.

The assumption that text representations can be modeled as Gaussian distribution seems similar to that in ProDA~\cite{lu2022prompt}.
However, in \cite{lu2022prompt}, the distribution is simply estimated using the statistics of multiple prompts such as $\boldsymbol{w}_{1:C}(\mathbf{P}) \sim \mathcal{N}(\boldsymbol{\mu}_{{w}_{1:C}(\mathbf{P})}, \boldsymbol{\sigma}_{{w}_{1:C}(\mathbf{P})}\mathbf{I})$.
In contrast, we model independent class-attribute distributions from multiple prompts considering the contextual information from image and text, and represent the class-specific prompt distribution as a mixture of distributions. 
Consequently, our method not only captures the diversity of visual representations but also alleviates the bias problem caused by aleatoric uncertainty, providing better generalization to the downstream tasks.
\vspace{-5pt}
\subsection{Training}
\vspace{-3pt}
\paragraph{Handling cross-modal uncertainty.}
Although \textit{a priori} distribution $p(\boldsymbol{w}_{1:C})$ of class representations can provide granular text representations, high uncertainty resulting from complex and diverse visual representation attenuates the reliability of predictive probabilities.
To mitigate the negative impact of uncertainty, we replace \equref{eq:pixel} with probabilistic pixel-text matching loss $\mathcal{L}_{prob}$, defined as:
\begin{equation}
    \mathcal{L}_{prob} = \frac{1}{(\prod_{c}^{C}\sigma(\boldsymbol{w}_{c})^2)^{1/C}}\mathcal{L}_{pixel} + \frac{1}{2C}\sum_{c}^{C}\log\sigma(\boldsymbol{w}_{c})^2,
\end{equation}
where the denominator $(\prod_{c}^{C}\sigma(\boldsymbol{w}_{c})^2)^{1/C}$ in the first term is to alleviate the penalty originating from uncertainty, and the second term reduces the high uncertainties.

\vspace{-12pt}
\paragraph{Total objectives.}
We employ additional KL divergence loss $\mathcal{L}_{KL}$ between each class distribution and Gaussian prior distribution $\mathcal{N}(0,I)$ to prevent the learned variance from collapsing to zero, inspired by \cite{chun2021pcme,park2022probabilistic}:
\begin{equation}
    \mathcal{L}_{KL} = \frac{1}{C}\sum_{c}^{C} \mathrm{KL}(p(z|\boldsymbol{w}_{c})\parallel \mathcal{N}(0,I)).
\end{equation}

Therefore, with the task-specific loss, $\mathcal{L}_{task}$ for dense prediction task, the overall objective of the proposed PPL is a weighted summation of all loss functions defined as:
\begin{equation}
    \mathcal{L}_{\mathbf{PPL}} = \mathcal{L}_{task} + \mathcal{L}_{prob} + \alpha\mathcal{L}_{div} + \beta\mathcal{L}_{KL},
    \label{eq:total}
 \end{equation}
where $\alpha$ and $\beta$ are hyper-parameters.
Note that $\mathcal{L}_{KL}$ and uncertainty regularization term in $\mathcal{L}_{prob}$ have opposite objectives: $\mathcal{L}_{KL}$ prevents $\sigma(\boldsymbol{w}_c)$ from collapsing to zero, while the uncertainty regularization term aims to reduce $\sigma(\boldsymbol{w}_c)$. 
The balance of these terms is controlled by $\beta$.

\vspace{-8pt}
\paragraph{Inference.} Given the learned prompts set $\{\mathbf{p}^{1}, ...,\mathbf{p}^{K}\}$, class representation $\boldsymbol{z}_{c}$ follows $\mathcal{N}(\mu(\boldsymbol{w}_{c}),\sigma(\boldsymbol{w}_{c}))$.
The prediction probability of the input image $\mathbf{x}$ is formulated by $\mathbb{E}_{\boldsymbol{w}_{1:C}}\left[p(y|\mathbf{x}_{i,j}, \boldsymbol{z}_{1:C})\right]$.
While computing the prediction probability is intractable over $p(\boldsymbol{w}_{1:C})$, it can be factorized via Monte-Carlo estimation, defined as:
\begin{equation}\label{eq:pred_prob}
\mathbb{E}_{\boldsymbol{w}_{1:C}}\left[p(y|\mathbf{x}_{i,j}, \boldsymbol{z}_{1:C})\right] \approx \frac{1}{N}\sum^{N}_{n}p(y|\mathbf{x}_{i,j}, \boldsymbol{z}_{1:C}^n).
\end{equation}
\vspace{-20pt}
\section{Experiments}
We present the experimental results to demonstrate the effectiveness of the proposed PPL. 
We conduct comparisons with the state-of-the-art methods on different dense prediction tasks: semantic segmentation, object detection, and instance segmentation.
Then, we provide the results of extensive ablation studies.
\subsection{Experimental Settings}
\vspace{-5pt}
For prompt learning, we set context length $L=8$ and initialize the context vectors using Gaussian noise.
\begin{table*}[t]
  \centering
  \caption{\textbf{Quantitative results of semantic segmentation on ADE20k dataset} under different pre-trained dataset and backbone network settings.
  (*: Results are directly taken from \cite{rao2022denseclip}.)}\vspace{-5pt}
  \resizebox{\textwidth}{!}{
  {\tiny
    \begin{tabular}{clcllcc}
    \toprule
    Backbone & Method & Pre-train & \multicolumn{1}{c}{mIoU-SS} & \multicolumn{1}{c}{mIoU-MS} & GFLOPs & Params(M) \\
    \specialrule{0.1pt}{0.5pt}{0.5pt}
    \multirowcell{8}{ResNet-50} & FCN~\cite{long2015fully}   & ImageNet & \multicolumn{1}{c}{36.1} & \multicolumn{1}{c}{38.1} & 793.6 & 49.6 \\
          & PSPNet~\cite{zhao2017pyramid} & ImageNet & \multicolumn{1}{c}{41.1} & \multicolumn{1}{c}{41.9} & 716.2 & 49.1 \\
          & Deeplab V3+~\cite{chen2018encoder} & ImageNet & \multicolumn{1}{c}{42.7} & \multicolumn{1}{c}{43.8} & 711.5 & 43.7 \\
          & UperNet~\cite{xiao2018unified} & ImageNet & \multicolumn{1}{c}{42.1} & \multicolumn{1}{c}{42.8} & 953.2 & 66.5 \\
          & Semantic FPN*~\cite{kirillov2019panoptic} & ImageNet & \multicolumn{1}{c}{38.6} & \multicolumn{1}{c}{40.6} & 227.1 & 31 \\
          & CLIP+Semantic FPN*~\cite{radford2021learning} & CLIP  & \multicolumn{1}{c}{39.6} & \multicolumn{1}{c}{41.6} & 248.8 & 31 \\
          & DenseCLIP+Semantic FPN*~\cite{rao2022denseclip} & CLIP  & \multicolumn{1}{c}{43.5} & \multicolumn{1}{c}{44.7} & 269.2 & 50.3 \\
          & \cellcolor[rgb]{ .886,  1,  .765}PPL + Semantic FPN & \cellcolor[rgb]{ .886,  1,  .765}CLIP & \multicolumn{1}{c}{\cellcolor[rgb]{ .886,  1,  .765}\textbf{44.7}} & \multicolumn{1}{c}{\cellcolor[rgb]{ .886,  1,  .765}\textbf{45.8}} & \cellcolor[rgb]{ .886,  1,  .765}421.4 & \cellcolor[rgb]{ .886,  1,  .765}51.8 \\[-0.15em]
    \specialrule{0.1pt}{0.5pt}{0.5pt}
    \multirowcell{8}{ResNet-101} & FCN~\cite{long2015fully}   & ImageNet & \multicolumn{1}{c}{39.9} & \multicolumn{1}{c}{41.4} & 1104.4 & 68.6 \\
          & PSPNet~\cite{zhao2017pyramid} & ImageNet & \multicolumn{1}{c}{43.6} & \multicolumn{1}{c}{44.4} & 1027.4 & 68.1 \\
          & Deeplab V3+~\cite{chen2018encoder} & ImageNet & \multicolumn{1}{c}{44.6} & \multicolumn{1}{c}{46.1} & 1022.7 & 62.7 \\
          & UperNet~\cite{xiao2018unified} & ImageNet & \multicolumn{1}{c}{43.8} & \multicolumn{1}{c}{44.8} & 1031  & 85.5 \\
          & Semantic FPN*~\cite{kirillov2019panoptic} & ImageNet & \multicolumn{1}{c}{40.4} & \multicolumn{1}{c}{42.3} & 304.9 & 50 \\
          & CLIP+Semantic FPN*~\cite{radford2021learning} & CLIP  & \multicolumn{1}{c}{42.7} & \multicolumn{1}{c}{44.3} & 236.6 & 50 \\
          & DenseCLIP+Semantic FPN*~\cite{rao2022denseclip} & CLIP  & \multicolumn{1}{c}{45.1} & \multicolumn{1}{c}{46.5} & 346.23 & 67.9 \\
          & \cellcolor[rgb]{ .886,  1,  .765}PPL + Semantic FPN & \cellcolor[rgb]{ .886,  1,  .765}CLIP & \multicolumn{1}{c}{\cellcolor[rgb]{ .886,  1,  .765}\textbf{46.4}} & \multicolumn{1}{c}{\cellcolor[rgb]{ .886,  1,  .765}\textbf{47.8}} & \cellcolor[rgb]{ .886,  1,  .765}496.9 & \cellcolor[rgb]{ .886,  1,  .765}69.4 \\[-0.15em]
    \specialrule{0.1pt}{0.5pt}{0.5pt}
    \multirowcell{6}{ViT-B} & SETR-MLA-DeiT~\cite{zheng2021rethinking} & ImageNet & \multicolumn{1}{c}{46.2} & \multicolumn{1}{c}{47.7} & -     & - \\
          & Semantic FPN*~\cite{kirillov2019panoptic} & ImageNet & \multicolumn{1}{c}{48.3} & \multicolumn{1}{c}{50.9} & 937.4 & 100.8 \\
          & Semantic FPN*~\cite{kirillov2019panoptic} & ImageNet-21K & \multicolumn{1}{c}{49.1} & \multicolumn{1}{c}{50.9} & 937.4 & 100.8 \\
          & CLIP+Semantic FPN*~\cite{radford2021learning} & CLIP  & \multicolumn{1}{c}{49.4} & \multicolumn{1}{c}{50.4} & 937.4 & 100.8 \\
          & DenseCLIP+Semantic FPN*~\cite{rao2022denseclip} & CLIP  & \multicolumn{1}{c}{50.6} & \multicolumn{1}{c}{50.3} & 933.1 & 105.3 \\
          & \cellcolor[rgb]{ .886,  1,  .765}PPL + Semantic FPN & \cellcolor[rgb]{ .886,  1,  .765}CLIP & \multicolumn{1}{c}{\cellcolor[rgb]{ .886,  1,  .765}\textbf{51.6}} & \multicolumn{1}{c}{\cellcolor[rgb]{ .886,  1,  .765}\textbf{51.8}} & \cellcolor[rgb]{ .886,  1,  .765}1072.1 & \cellcolor[rgb]{ .886,  1,  .765}106.9 \\[-0.25em]
    \bottomrule
    \end{tabular}
    }}
    \vspace{-15pt}
  \label{tab:semseg}%
\end{table*}%
\begin{table}[t!]
  \centering
  \caption{\textbf{Semantic segmentation performance on ADE20k} of ProDA~\cite{lu2022prompt} and the proposed method.}\vspace{-5pt}
  \resizebox{\columnwidth}{!}{
    \begin{tabular}{lccc}
    \toprule
    Model & mIoU-SS & GFLOPs & Params(M) \\
    \midrule
    ProDA+Semantic FPN~\cite{lu2022prompt} &  42.6  & 1379  & 46.4 \\
    PPL+Semantic FPN  & \textbf{44.7}  &  421  & 51.8 \\
    \bottomrule\vspace{-25pt}
    \end{tabular}}
  \label{tab:proda_semseg}%
\end{table}%
We freeze the text encoder to conserve the pre-trained language knowledge.
The visual-context probabilistic decoder is composed of a transformer module with 5 layers and 1 variance prediction module.
We train our network with AdamW optimizer.
In the comparison experiments, we set the number of attributes $K=3$, and sampled the embeddings $N=15$ fairly and evenly from each of the $K$ distributions, and the KL-divergence hyperparameter $\beta=10^{-5}$ throughout the experiments.
We also include FLOPs and the number of parameters for fair comparisons.
Additional settings and implementation details for each experiment are presented in each subsection.
\vspace{-5pt}
\subsection{Semantic  Segmentation}
\vspace{-5pt}
\paragraph{Settings.}\label{para:semsetup} We evaluate the semantic segmentation performance of PPL on ADE20k~\cite{zhou2019semantic}, which contains 20k training and 2k validation images with 150 categories.
Following the common protocol in ~\cite{huang2019ccnet,xiao2018unified}, we evaluate the performance on the validation set, using mIoU scores measured in single scale (mIoU-SS) and multiple scales (mIoU-MS).
We adopt the Semantic FPN~\cite{kirillov2019panoptic} framework, with different encoders: ResNet-50 (RN50), ResNet-101 (RN101)~\cite{he2016deep}, and ViT-B~\cite{dosovitskiy2020vit}.
The network is trained for 127 epochs with a batch size of 32 and a learning rate of $10^{-4}$.
\vspace{-12pt}
\paragraph{Results.}
\tabref{tab:semseg} shows the semantic segmentation performance of different methods on ADE20k dataset~\cite{zhou2019semantic}.
\begin{figure}[t!]
    \centering
\begin{subfigure}[t]{0.24\columnwidth}
    \includegraphics[width=\textwidth]
    {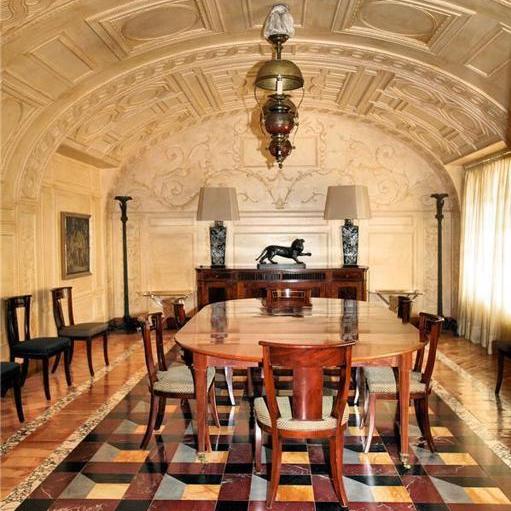}
    \includegraphics[width=\textwidth]
    {figures/fig6/i612.jpg}
    \includegraphics[width=\textwidth]
    {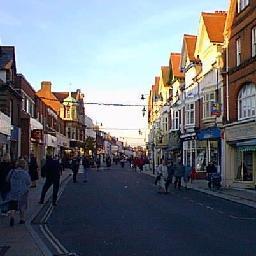}
    \includegraphics[width=\textwidth]
    {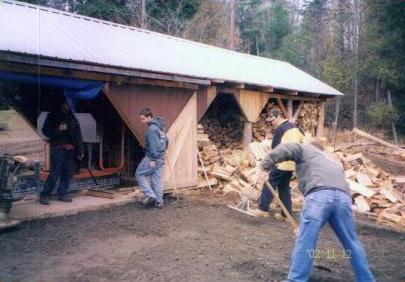}
    \caption{Image}
\end{subfigure}
\begin{subfigure}[t]{0.24\columnwidth}
    \includegraphics[width=\textwidth]
    {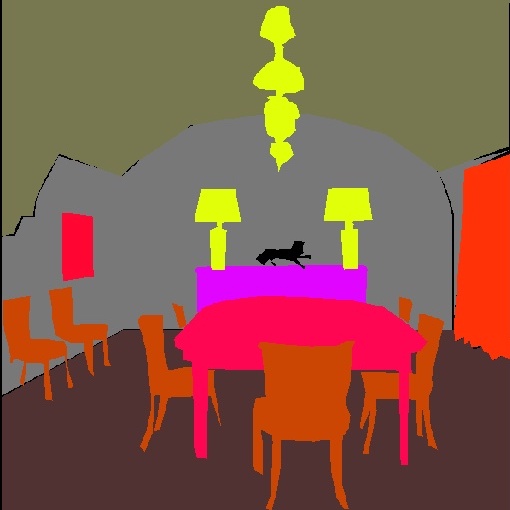}
    \includegraphics[width=\textwidth]
    {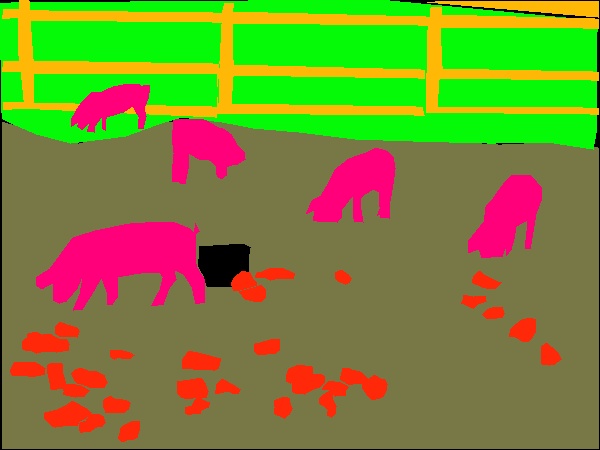}
    \includegraphics[width=\textwidth]
    {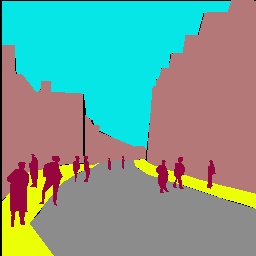}
    \includegraphics[width=\textwidth]
    {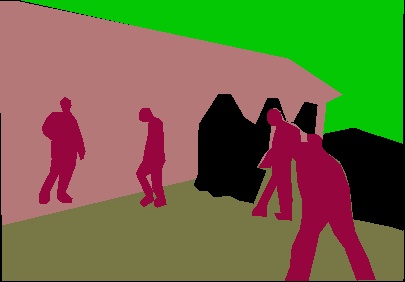}
    \caption{Ground-truth}
\end{subfigure}
\begin{subfigure}[t]{0.24\columnwidth}
    \includegraphics[width=\textwidth]
    {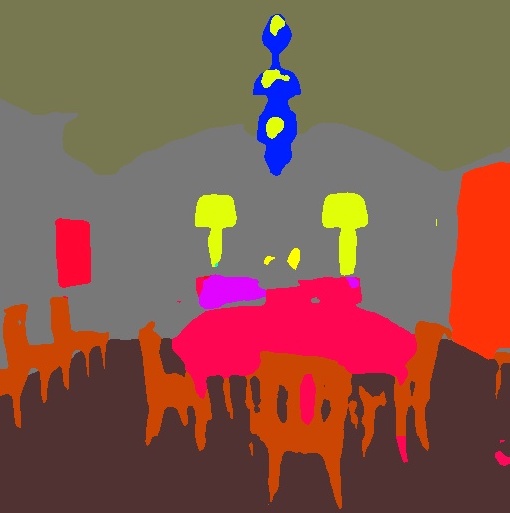}
    \includegraphics[width=\textwidth]
    {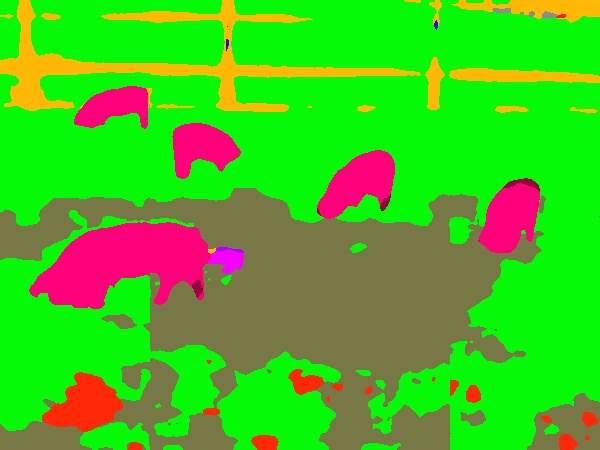}
    \includegraphics[width=\textwidth]
    {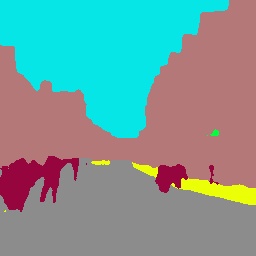}
    \includegraphics[width=\textwidth]
    {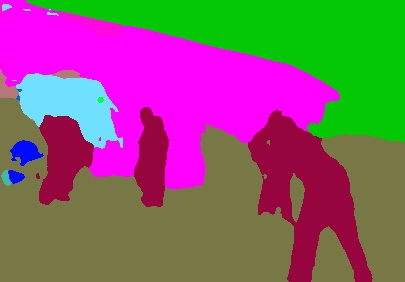}
    \caption{DenseCLIP}
\end{subfigure}
\begin{subfigure}[t]{0.24\columnwidth}
    \includegraphics[width=\textwidth]
    {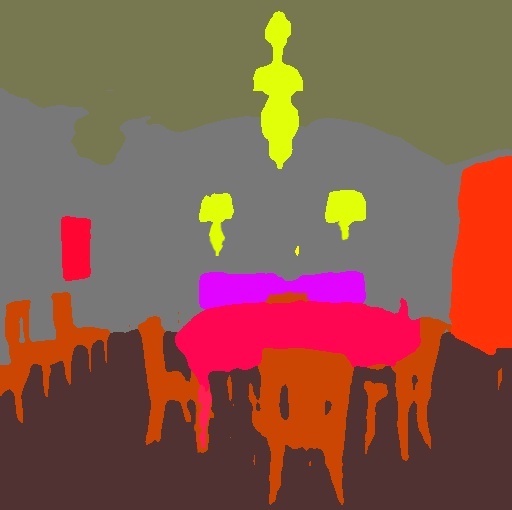}
    \includegraphics[width=\textwidth]
    {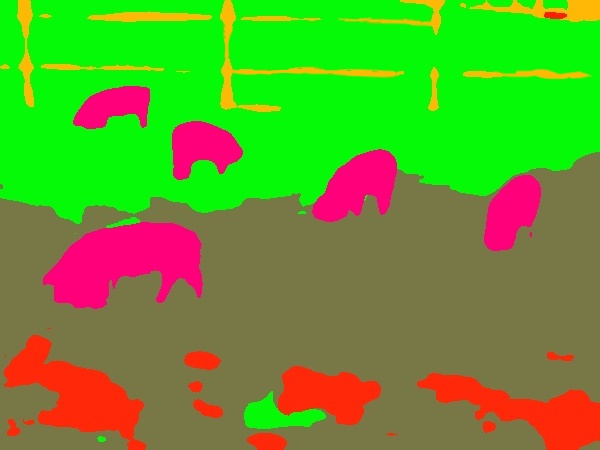}
    \includegraphics[width=\textwidth]
    {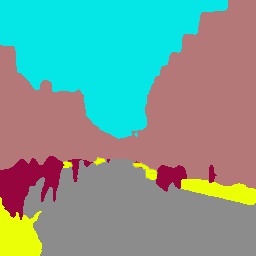}
    \includegraphics[width=\textwidth]
    {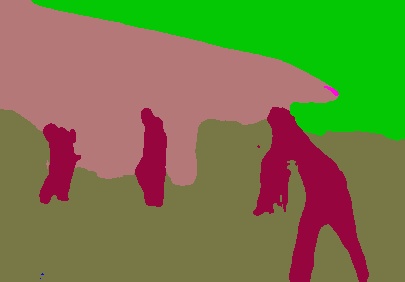}
    \caption{PPL}
\end{subfigure}
\vspace{-8pt}\caption{\textbf{Qualitative results} for semantic segmentation on ADE20k based on DenseCLIP and our proposed PPL.}
\vspace{-25pt}
\label{fig:qualitative}
\end{figure}
The comparison between DenseCLIP and our method shows that our probabilistic approach outperforms the deterministic approach by 1.2\% for RN50, 1.3\% for RN101, and 1\% for ViT-B backbones, in terms of mIoU-SS.
Furthermore, compared to the vanilla CLIP-based segmentation networks \cite{rao2022denseclip}, our approach results in higher mIoU-MS by 5.1\%, 3.7\%, and 2.2\% for RN50, RN101, and ViT-B encoders, respectively.
\begin{table*}[t]
  \centering
  \caption{\textbf{Quantitative results of object detection and instance segmentation on COCO} with different backbone networks.(*: Results are directly taken from \cite{rao2022denseclip}.)}\vspace{-5pt}
  \resizebox{\textwidth}{!}{
    \begin{tabular}{lcccccccc|cccccc}
    \toprule
    \multirowcell{2}{Model} & \multirowcell{2}{GFLOPs} & \multirowcell{2}{Params(M)} & \multicolumn{6}{c|}{Object Detection}         & \multicolumn{6}{c}{Instance Segmentation} \\
\cmidrule{4-15}          &       &       & $AP^{b}$ & $AP^{b}_{50}$ & $AP^{b}_{75}$ & $AP^{b}_{S}$ & $AP^{b}_{M}$ & $AP^{b}_{L}$ & $AP^{m}$ & $AP^{m}_{50}$ & $AP^{m}_{75}$ & $AP^{m}_{S}$ & $AP^{m}_{M}$ & $AP^{m}_{L}$ \\
    \specialrule{0.1pt}{0.5pt}{0.5pt}
    RN50-IN1K~\cite{he2016deep} & 275   & 44    & 38.2  & 58.8  & 41.4  & 21.9  & 40.9  & 49.5  & 34.7  & 55.7  & 37.2  & 18.3  & 37.4  & 47.2 \\
    RN50-CLIP*~\cite{radford2021learning} & 301   & 44    & 39.3  & 61.3  & 42.7  & 24.6  & 42.6  & 50.1  & 36.8  & 58.5  & 39.2  & 18.6  & 39.9  & 51.8 \\
    RN50-DenseCLIP*~\cite{rao2022denseclip} & 327   & 67    & 40.2  & 63.2  & 43.9  & 26.3  & 44.2  & 51    & 37.6  & 60.2  & 39.8  & 20.8  & 40.7  & 53.7 \\
    \rowcolor[rgb]{ .886,  1,  .765} RN50-PPL & 368     & 70     & \textbf{41.0}     & \textbf{64.5}     & \textbf{45.2}    & \textbf{27.0}     & \textbf{45.1}     & \textbf{51.8}    & \textbf{38.3}     & \textbf{60.9}     & \textbf{41.1}     & \textbf{21.1}     & \textbf{41.3}     & \textbf{53.9} \\
    \specialrule{0.1pt}{0.5pt}{0.5pt}
    RN101-IN1K~\cite{he2016deep} & 351   & 63    & 40.0    & 60.5  & 44.0    & 22.6  & 44    & 52.6  & 36.1  & 57.5  & 38.6  & 18.8  & 39.7  & 49.5 \\
    RN101-CLIP*~\cite{radford2021learning} & 377   & 63    & 42.2  & 64.2  & 46.5  & 26.4  & 46.1  & 54.0    & 38.9  & 61.4  & 41.8  & 20.5  & 42.3  & 55.1 \\
    RN101-DenseCLIP*~\cite{rao2022denseclip} & 399   & 84    & 42.6  & 65.1  & 46.5  & 27.7  & 46.5  & 54.2  & 39.6  & 62.4  & 42.4  & 21.4  & 43.0    & 56.2 \\
    \rowcolor[rgb]{ .886,  1,  .765} RN101-PPL & 444     & 87     & \textbf{43.2}     & \textbf{66.0}   & \textbf{47.0}     & \textbf{28.1}     & \textbf{47.1}     & \textbf{54.8}     & \textbf{40.4}     & \textbf{63.3}     & \textbf{43.2}     & \textbf{21.8}     & \textbf{43.7}    & \textbf{57.0} \\[-0.25em]
    \bottomrule
    \end{tabular}}
    \vspace{-15pt}
  \label{tab:obis}%
\end{table*}%
\begin{table}[t]
  \centering
  \caption{\textbf{Ablation studies} on component of objective functions.}\vspace{-5pt}
  \resizebox{0.9\columnwidth}{!}{
  {\small	
    \begin{tabular}{p{1.5cm}p{1.5cm}p{1.5cm}p{1cm}p{1cm}}
    \toprule
    \multicolumn{1}{c}{$\mathcal{L}_{div}$} & \multicolumn{1}{c}{$\mathcal{L}_{KL}$} & \multicolumn{1}{c}{$\mathcal{L}_{prob}$} & \multicolumn{1}{c}{mIoU} & \multicolumn{1}{c}{$\Delta$} \\
    \midrule
          &       &       & \multicolumn{1}{c}{42.0}    & \multicolumn{1}{c}{0} \\
    \multicolumn{1}{c}{\checkmark}  &       &       & \multicolumn{1}{c}{42.4}  & \multicolumn{1}{c}{+0.4} \\
          & \multicolumn{1}{c}{\checkmark}  &  & \multicolumn{1}{c}{43.6}  & \multicolumn{1}{c}{+1.6} \\
    \multicolumn{1}{c}{\checkmark}  &  \multicolumn{1}{c}{\checkmark}     &   & \multicolumn{1}{c}{44.3}  & \multicolumn{1}{c}{+2.3} \\
    \multicolumn{1}{c}{\checkmark}  & \multicolumn{1}{c}{\checkmark}  & \multicolumn{1}{c}{\checkmark}  & \multicolumn{1}{c}{44.7}  & \multicolumn{1}{c}{+2.7} \\
    \bottomrule
    \end{tabular}}}
    \vspace{-15pt}
  \label{tab:abl_loss}%
\end{table}%
We provide qualitative results of DenseCLIP~\cite{rao2022denseclip} and the proposed method in Fig. \ref{fig:qualitative}.
We observe that our method tends to capture fine-detailed objects and segment the correct labels compared to DenseCLIP.
Furthermore, since our method leverage various visual-context knowledge, we can reduce error in the ambiguous region to classify, and this is shown through \figref{fig:qualitative}.
\vspace{-12pt}
\paragraph{Comparison with ProDA}
We further compare our approach with the existing probabilistic approach ProDA~\cite{lu2022prompt}.
For a fair comparison, both methods use RN-50 backbone, the number of sampled prompts $N_{ProDA}=16$, $N_{PPL}=15$.
The quantitative result is presented in \tabref{tab:proda_semseg}. 
We observe higher mIoU-SS of the proposed method.
Since the proposed PPL estimates the class-specific attribute distribution by considering the visual-text relationship, it is a more suitable segmentation.
On the other hand, ProDA uses only textual information to model the distribution, so it tends to fail to capture the diverse and complex components.
\vspace{-5pt}
\subsection{Object Detection and Instance Segmentation}
\vspace{-3pt}
\paragraph{Settings.}
We evaluate the proposed PPL in object detection and instance segmentation on COCO dataset~\cite{lin2014microsoft}, which consists of 118k training samples and 5k validation images.
We conduct experiments on our proposed method using Mask-RCNN architecture~\cite{he2017mask}.
We report results using standard Average Precision metric measured using bounding box ($AP^b$) and segmentation mask ($AP^m$) with IoU$=0.5/0.75$, and object sizes.
We use RN50 and RN101 as backbone networks.
The network is trained for 12 epochs with a batch size of 16 and $2\times10^{-4}$.
\vspace{-12pt}
\paragraph{Results.}
We summarized the experimental results in \tabref{tab:obis}.
We observe the VLM methods outperform the conventional ImageNet-1K (IN1K) pre-trained model. 
The proposed PPL exploits multiple text representations to provide diverse visual-language knowledge to generate plausible results, and outperforms both Vanilla CLIP \cite{zhou2022learning} and DenseCLIP \cite{rao2022denseclip} in both object detection and instance segmentation. 
\begin{table}[t]
  \centering
  \caption{\textbf{Ablation studies} on hyperparameters.
  The number of attributes $K$, KL-divergence hyperparameter $\beta$, and the number of sampled representations $N$.}\vspace{-5pt}
  \resizebox{0.9\columnwidth}{!}{
    \begin{tabular}{p{2.5cm}p{1cm}p{1cm}p{1cm}p{1cm}}
    \specialrule{0.1pt}{0.5pt}{0.5pt}
    \rowcolor[rgb]{ .949,  .949,  .949}\multicolumn{5}{l}{\textbf{Number of attributes} $(N=15, \beta=10^{-5})$} \\
    Parameter $K$ & $1$   & $3$   & $5$   & $7$ \\
    mIoU  & 43.8     &\textbf{44.7}     & 44.2     & 44.0 \\
    \specialrule{0.1pt}{0.5pt}{0.5pt}
    \rowcolor[rgb]{ .949,  .949,  .949}\multicolumn{5}{l}{\textbf{KL-divergence hyperparameter} $(N=15, K=3)$} \\
    Parameter $\beta$ & $10^{-7}$ & $10^{-6}$ & $10^{-5}$ & $10^{-4}$ \\
    mIoU  & 42.4     & 42.8     & \textbf{44.7}    & 44.1 \\
    \specialrule{0.1pt}{0.5pt}{0.5pt}
    \rowcolor[rgb]{ .949,  .949,  .949}\multicolumn{5}{l}{\textbf{Number of sampled representations} $(\beta=10^{-5}, K=3)$} \\
    Parameter $N$ & $5$   & $10$  & $15$  & $20$ \\
    mIoU  & 43.9     & 44.4     & 44.7     & \textbf{44.8} \\
    \bottomrule
    \end{tabular}}
    \vspace{-15pt}
  \label{tab:abbl_param}%
\end{table}%
Showing consistent improvements across the sub-measures, we confirm that the diverse expressions of visual-context from the PPL help the network to generate more accurate predictions across different scales with precise localization.
\vspace{-5pt}
\subsection{Ablation Study and Analysis}
\vspace{-4pt}
To further analyze and validate the components of our method, we conduct ablation study experiments on semantic segmentation.
In addition, to show the impact of multiple attributes prompts,  we visualize the activation maps and analyze them in detail.
\vspace{-12pt}
\paragraph{Contribution of objectives.}
We conduct experiments to observe the effect of each loss function in (\ref{eq:total}).
To this end, we train networks with different objective functions and present the results in \tabref{tab:abl_loss}.
We first observe 42.0 mIoU of a baseline network trained using only $\mathcal{L}_{task}$ and $\mathcal{L}_{pixel}$ in (\ref{eq:pixel}).
Applying each of the diversity loss $\mathcal{L}_{div}$ and KL divergence loss $\mathcal{L}_{KL}$ encourages the diversity in probabilistic distribution, resulting in improved performance.
The performance is further improved when they are applied simultaneously. 
Finally, replacing $\mathcal{L}_{pixel}$ with $\mathcal{L}_{prob}$ regularizes high uncertainties and results in further improved performance, achieving 44.7 mIoU.
The analysis of $\mathcal{L}_{prob}$ is described in the last paragraph of \secref{sec:abbl_uncert}.
\vspace{-12pt}
\paragraph{Number of attribute prompts.}
We compare the semantic segmentation performance with respect to the number of attributes prompts $K$.
As shown in the first block of \tabref{tab:abbl_param}, $K=3$ notably outperforms $K=1$, which justifies the introduction of multiple attribute representations.
\begin{figure}[t!]
		\centering
		\begin{subfigure}[t]{0.525\columnwidth}
    \includegraphics[width=\textwidth]
    {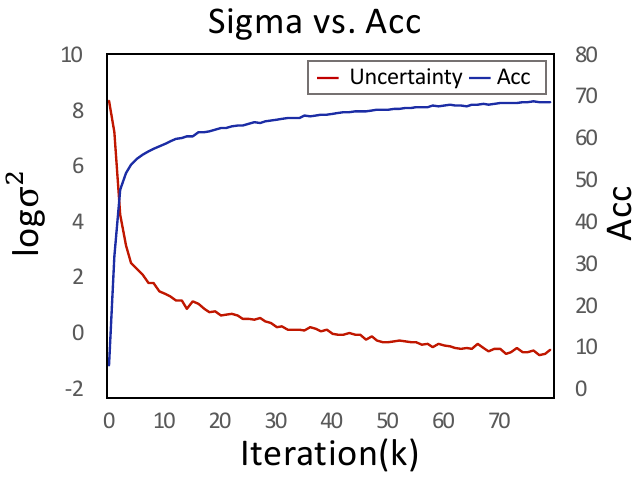}
\end{subfigure}
\begin{subfigure}[t]{0.465\columnwidth}
    \includegraphics[width=\textwidth]
    {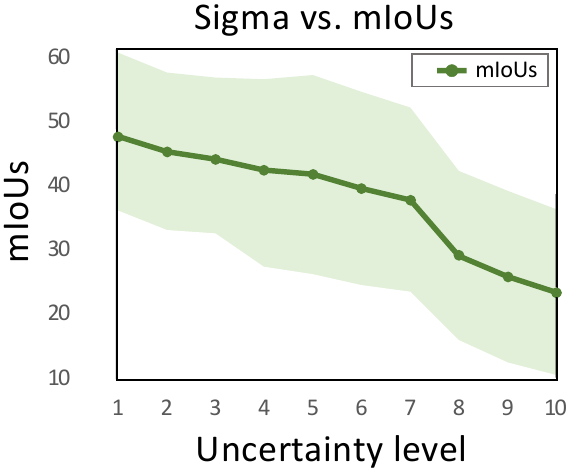}
\end{subfigure}

	    \caption{\textbf{Illustration of uncertainty analysis} on semantic segmentation. (Left) uncertainty versus accuracy during training, (Right) performance versus uncertainty level for the test set.}
            \vspace{-20pt}
         \label{fig:abbl_uncert}
\end{figure}
However, we find that the performance is rather degraded for $K>3$ since many attributes are likely to provide redundant information, which degrades performance.

\vspace{-12pt}
\paragraph{KL-divergence hyperparameters.}
To investigate the effect of the KL-divergence hyperparameter $\beta$ in \equref{eq:total}, we include the experimental result in the second block of \tabref{tab:abbl_param}.
In general, the variance of a mixed distribution follows the unit variance as $\beta$ increases, reducing the discriminability of distributions.
Conversely, if $\beta$ is too small (\eg $10^{-6}$), variance converges to zero and $\mathcal{L}_{prob}$ diverges.
In summary, $\beta$ controls the range of use of the visual-context, and also adjusts $\mathcal{L}_{KL}$ and $\mathcal{L}_{prob}$, which operate opposite to each other, learn in a balanced way.
We find that our model achieves the best performance with $\beta=10^{-5}$. 
\vspace{-14pt}
\paragraph{Number of sampled representations.}
To analyze the effect of the number of sampled representations $N$, we conduct experiments with 5, 10, 15, and 20 samples.
The results are summarized in the third block of \tabref{tab:abbl_param}.
We can observe that the performance increases as the number of sampled representations increases.
For example, there was a $0.8\%$ improvement in performance when $N=15$ compared to when $N=5$.
However, due to the increase in computational cost as the number of sampled representations increases, we fixed $N=15$ to balance between the computational cost and performance.
\vspace{-14pt}
\paragraph{Uncertainty vs. Performance.}
To analyze the correlation between uncertainty and the prediction probability (\ref{eq:pred_prob}), we measured the uncertainty of text representations and report the performance in terms of segmentation accuracy. according to the training iterations.
We define the uncertainty as the geometric mean of the variance $\boldsymbol{\sigma}(\boldsymbol{w}_{1:C})$ given the input image.
As shown in the left plot in Fig.~\ref{fig:abbl_uncert}, the uncertainty is minimized and performance increases as the learning progress. 
Concretely, the network learns to focus on useful visual-contexts and neglect the redundant ones.

In addition, we divide the uncertainty value into 10 bins and measure the performance according to its level.
We observed a negative correlation between uncertainty and performance.
As shown in the right plot in Fig.~\ref{fig:abbl_uncert}, we observe that high uncertainty results in unreliable predictions, thus we suppress the high uncertainties via \eqref{eq:pred_prob} to generate the improved results.
\begin{figure}
\centering
\begin{subfigure}[t]{0.24\columnwidth}
    \makebox[0pt][r]{\makebox[15pt]{\raisebox{25pt}{\rotatebox[origin=c]{90}{\textit{lamp}}}}}%
    \includegraphics[width=\textwidth]
    {figures/fig6/i312.jpg}
    \makebox[0pt][r]{\makebox[15pt]{\raisebox{20pt}{\rotatebox[origin=c]{90}{\textit{person}}}}}%
    \includegraphics[width=\textwidth]
    {figures/fig5/4/i988.jpg}
    \makebox[0pt][r]{\makebox[15pt]{\raisebox{20pt}{\rotatebox[origin=c]{90}{\textit{sofa}}}}}%
    \includegraphics[width=\textwidth]
    {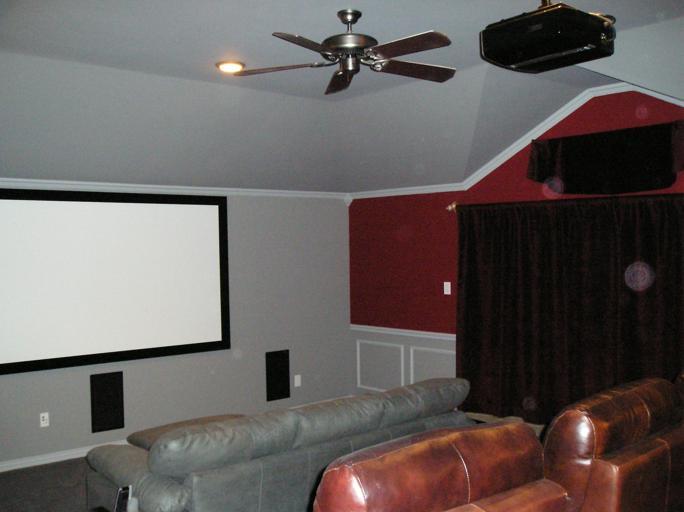}
    \makebox[0pt][r]{\makebox[15pt]{\raisebox{20pt}{\rotatebox[origin=c]{90}{\textit{food}}}}}%
    \includegraphics[width=\textwidth]
    {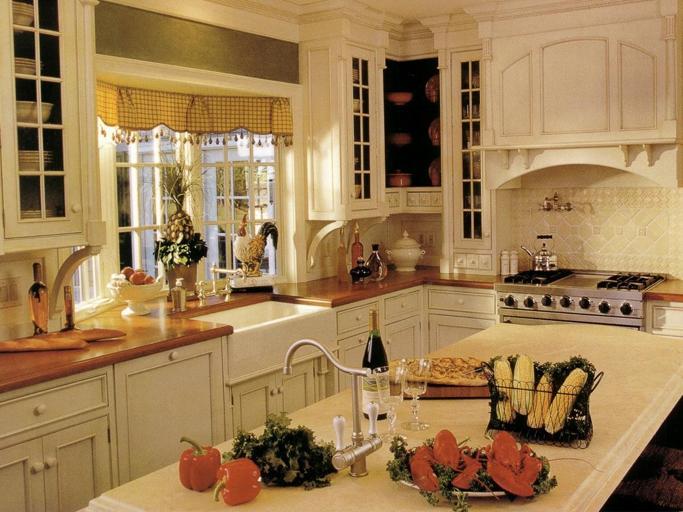}
    \caption{Image}
\end{subfigure}
\begin{subfigure}[t]{0.24\columnwidth}
    \includegraphics[width=\textwidth]  
    {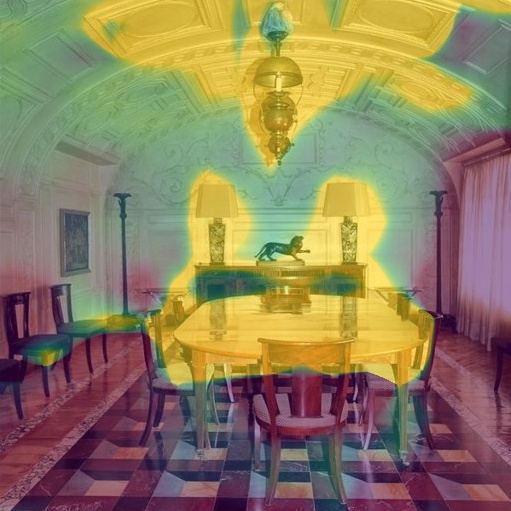}
    \includegraphics[width=\textwidth]
    {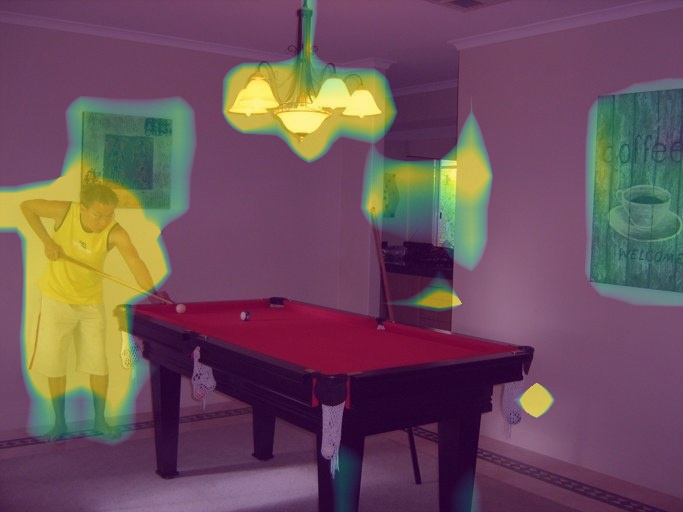}
    \includegraphics[width=\textwidth]
    {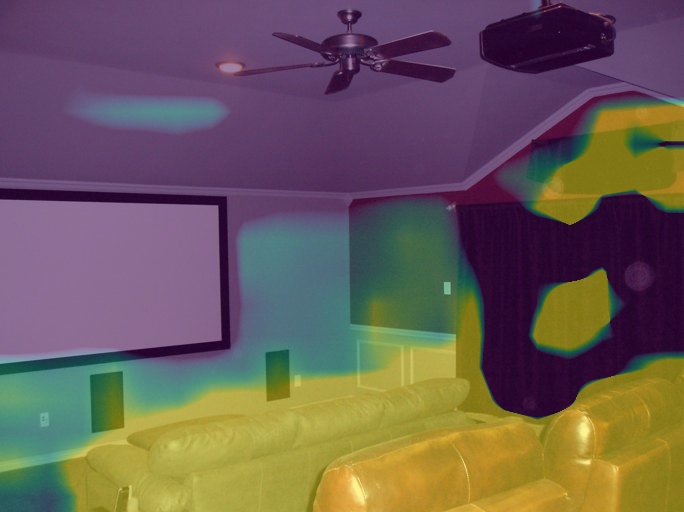}
    \includegraphics[width=\textwidth]
    {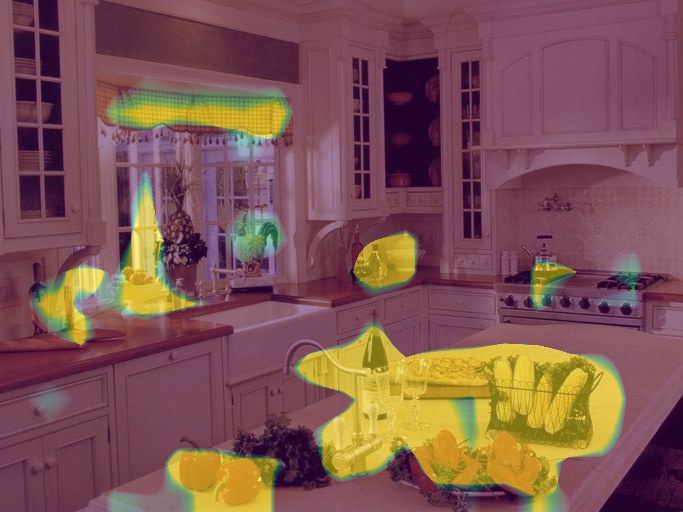}
    \caption{\texttt{Samples 1}}
\end{subfigure}
\begin{subfigure}[t]{0.24\columnwidth}
    \includegraphics[width=\textwidth]  
    {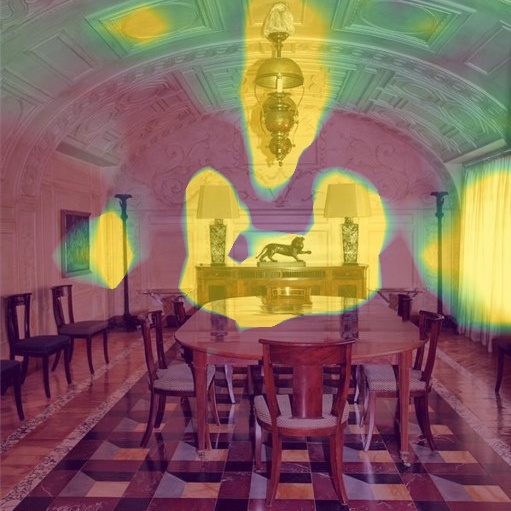}
    \includegraphics[width=\textwidth]
    {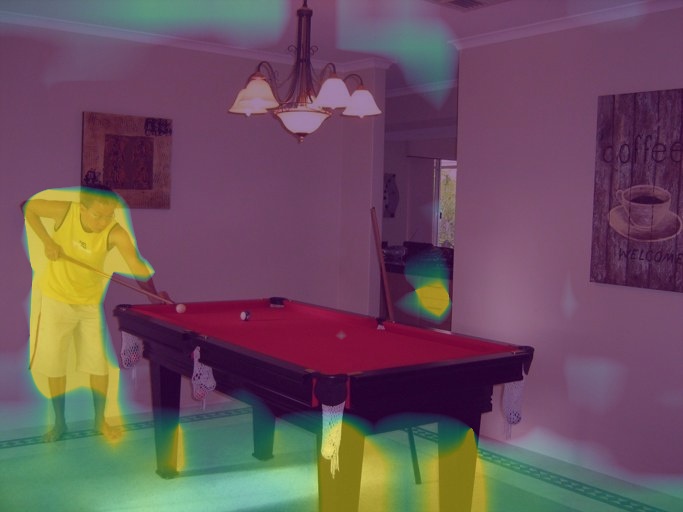}
    \includegraphics[width=\textwidth]
    {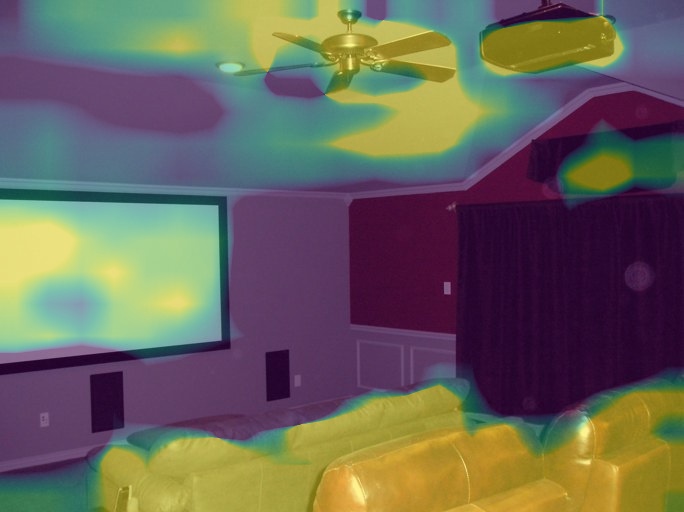}
    \includegraphics[width=\textwidth]
    {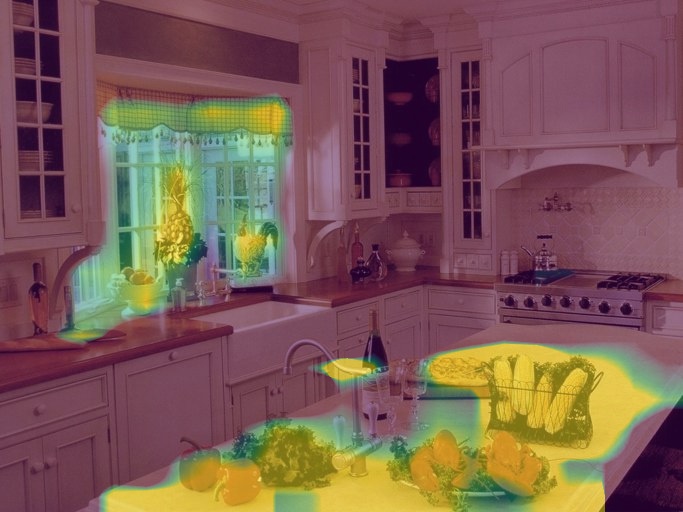}
    \caption{\texttt{Samples 2}}
\end{subfigure}
\begin{subfigure}[t]{0.24\columnwidth}
    \includegraphics[width=\textwidth]  
    {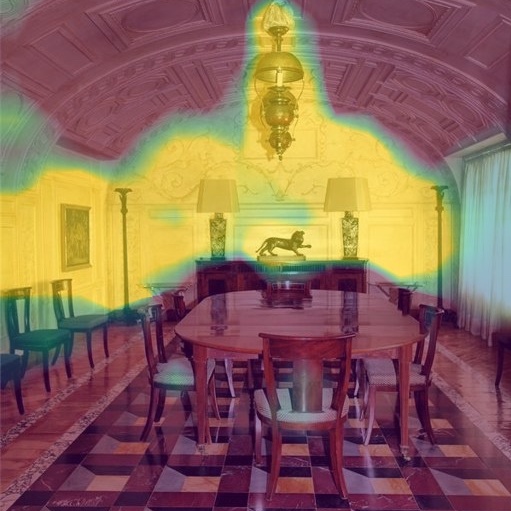}
    \includegraphics[width=\textwidth]
    {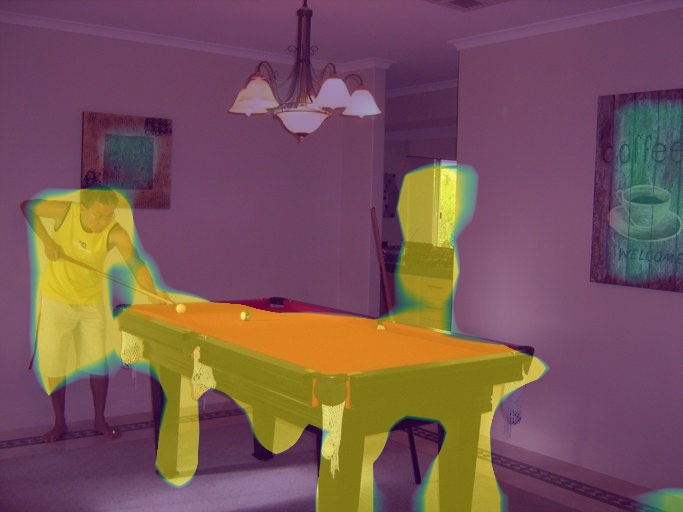}
    \includegraphics[width=\textwidth]
    {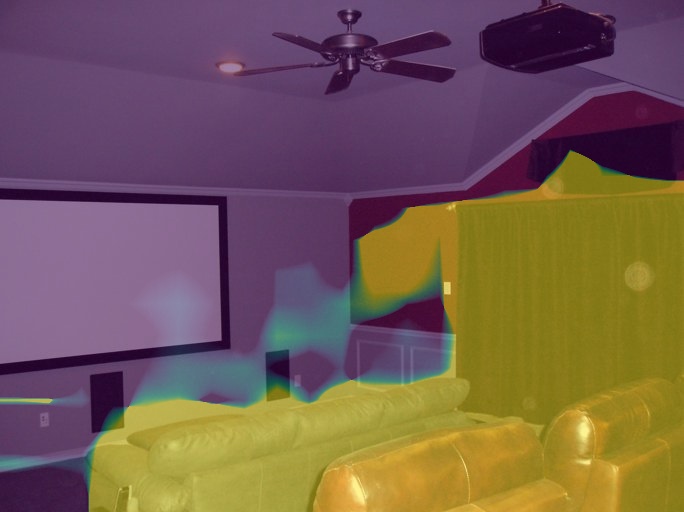}
    \includegraphics[width=\textwidth]
    {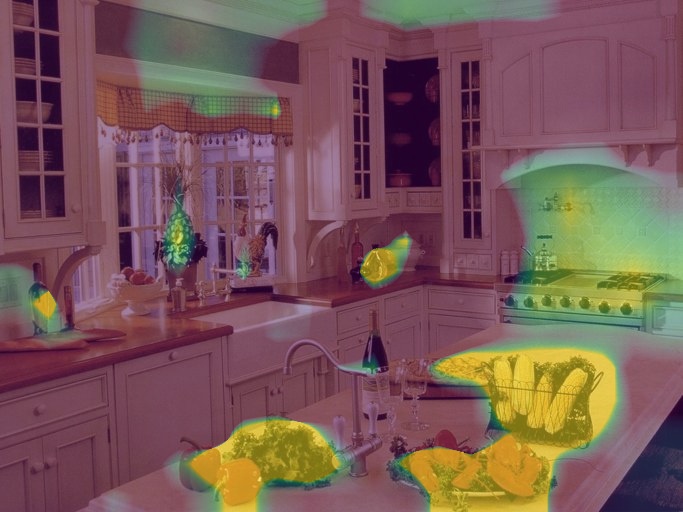}
    \caption{\texttt{Samples 3}}
\end{subfigure}
\vspace{-4pt}
\caption{\textbf{\textbf{Visualization of activation maps}}. We report the activation maps of each attribute distribution (mean) of different classes indicated on the left side, with $K=3$ on the ADE20k dataset. Different samples from MoG attend to different attributes of the object.}
\label{fig:act}
\vspace{-15pt}
\end{figure}
\vspace{-12pt}
\paragraph{Visualization}
To better understand the advantage of probabilistic embedding, we demonstrate some visualization examples of the activation maps derived from each sample text representation in \figref{fig:act}.
We observed that the attended region by sampled representations associated with different contexts changes according to the given context.
This means that these multiple prompts can efficiently represent objects of various shapes, sizes, and colors, and consequently, our method is suitable for dense prediction tasks.
\label{sec:abbl_uncert}
\vspace{-14pt}
\section{Conclusion}
\vspace{-6pt}
In this paper, we presented a probabilistic prompt learning (\textbf{PPL}) for dense prediction.
It aims to extract diverse text representations to fully exploit the knowledge from VLM.
Leveraging the visual-context information,  class-specific probabilistic text distribution is defined, from where diverse text representations are sampled to guide the dense prediction tasks.
In addition, we learn the optimal distribution by suppressing the high uncertainty from the complex visual-context via the probabilistic pixel-text matching loss. 
The experimental results show that the proposed method achieves significantly improved performance compared to the previous method in semantic segmentation, object detection, and instance segmentation, demonstrating the potential extension of our method to comprehensive multi-modal scene understanding tasks.
\vspace{-12pt}
\paragraph{Acknowledgement.} This research was supported by the Yonsei Signature Research Cluster Program of 2022 (2022-22-0002) and the KIST Institutional Program (Project No.2E31051-21-203). .

\end{document}

{\small
\bibliographystyle{ieee_fullname}
\bibliography{main}
}
\clearpage
\setcounter{figure}{0}


\appendix

\section{Appendix}
In this document, we include supplementary materials for PPL.
Firstly, we provide methodological details on PPL (Sec.~\ref{sec:uncert}) and Pseudo-code of PPL (Sec.~\ref{sec:alg}).
Furthermore, we provide the additional qualitative results for dense prediction tasks (Sec.~\ref{sec:add1}).
\vspace{-6pt}
\section{Uncertainty on PPL}
In this section, we provide how to measure uncertainty of text representations with visual context.
In addition, we report the cause of uncertainty.
\label{sec:uncert}
\subsection{Uncertainty estimation}
In PPL, the whole distribution of each class of given input image is estimated as a Mixture of Gaussian (MoG) with $K$-attribute prompts.
To compute uncertainty of images, we describe the computation of mean and variance of each class of image.
The PDF of a MoG of each class is represented by the average PDF of its attribute distributions of image given.
\begin{equation}\tag{15}
    f_{\boldsymbol{w}_c}(z)
    =\frac{1}{K}\sum f_{{w}^{k}_{c}}(z).
\end{equation}
Then, the mean of the MoG is formulated follow as:
\begin{align}
    \mu({\boldsymbol{w}_c})\nonumber
    &=\int zf_{\boldsymbol{w}_c}(z)dz\\\nonumber
    &=\frac{1}{K}\sum\int zf_{{w}^{k}_{c}}(z)dz\\\tag{16}
    &=\frac{1}{K}\sum\mu_c^k.\\\nonumber
    \vspace{-10pt}
\end{align}
The standard deviation $\sigma(\boldsymbol{w}_c)^2$ is derived as follow:
\begin{align}
    \sigma({\boldsymbol{w}_c})^2\nonumber
    &=\int z^2f_{\boldsymbol{w}_c}(z)dz - \mu({\boldsymbol{w}_c})^2\\\nonumber
    &=\frac{1}{K}\sum\int z^2f_{{w}^{k}_{c}}(z)dz - \mu({\boldsymbol{w}_c})^2\\\tag{17}
    &=\frac{1}{K}\sum((\mu_c^k)^2+(\sigma_c^k)^2) - (\frac{1}{K}\sum\mu_c^k)^2.\\\nonumber
    \vspace{-10pt}
\end{align}
We define the geometric mean of the variance $\sigma({\boldsymbol{w}_c})$ of each class $c$ is used as uncertainty of its class.
Finally, we formulate the total uncertainty of image is derived as follow:
\begin{equation}
    \bar{\sigma}(\boldsymbol{w}_{1:C})_{\mathrm{G}} = \prod(\sigma(\boldsymbol{w}_{c}))^{1/c}.\tag{18}
\vspace{-10pt}
\end{equation}
\begin{figure}[t]
    \centering
    \includegraphics[width=0.95\linewidth]{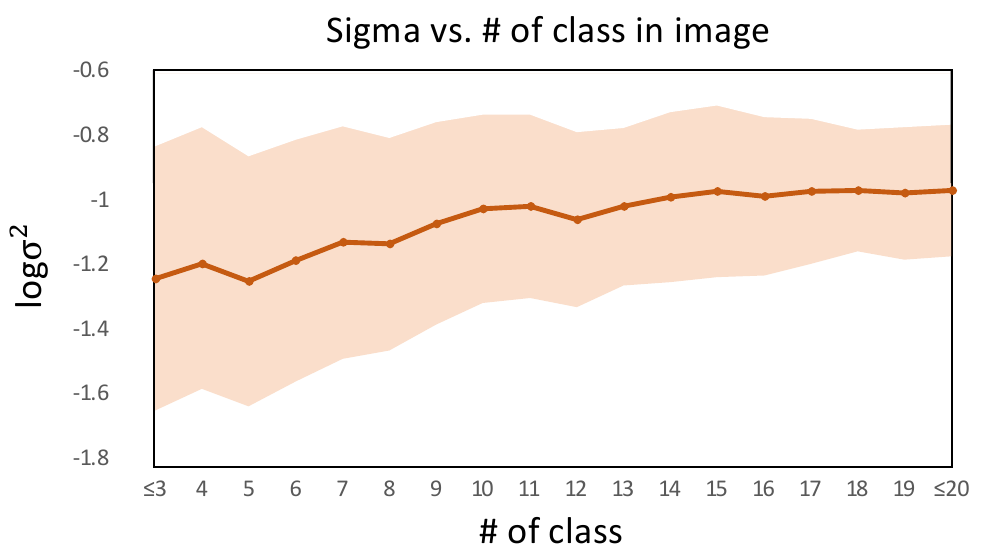}
    \vspace{-5pt}
    \caption{We measure the uncertainty of images on ADE20k~\cite{zhou2019semantic} dataset according to the number of classes included in the image.
} 
\vspace{-10pt}
\label{fig:suppl1}
\end{figure}

\subsection{Uncertainty Analysis}
To better understand uncertainty, we provide a brief analysis of the causes of uncertainty.
Although it is impossible to estimate all causes, we studied the correlation between the number of classes in an image and uncertainty.
As shown in Fig.~\ref{fig:suppl1}, the number of classes in image have positive relationship with uncertainty. 
Based on this results, the predicted uncertainty can be used to remove ambiguous visual-context and leverage the useful context in image.
\section{Algorithm}
\label{sec:alg}
\vspace{-4pt}
\begin{algorithm}
    \caption{Pseudo-code of PPL Training.}
    \label{al_ppl}
    \LinesNumbered
    \KwRequire{The pre-trained CLIP text encoder $\mathcal{G}$,image encoder $\mathcal{F}$, and visual-context probabilistic decoder $\mathcal{M}$}
    \KwRequire{Class descriptions $\mathbf{t}_{1:C}(\cdot)$ and randomly initialized prompts set $\mathbf{P}=\left[\mathbf{p}^1,...\mathbf{p}^N\right]$}
    \For{$t$ to $T$ do}{
        Draw a mini-batch ($\mathbf{x}$, $y$).\\
        Compute $v=\mathcal{F}(\mathbf{x})$ and $\boldsymbol{w}_{1:C}=\mathcal{G}(\mathbf{t}_{1:C}(\mathbf{P}))$\\
        Let $\boldsymbol{w}_{c}=\left[w_c^1,...,w_c^K\right] $\\
        Compute $\mathcal{L}_{div}$ according to Eq. (5)\\
        Compute $\boldsymbol{\sigma}_c^k=\mathcal{M}(w_{c}^k, v)$\\
        Compute $p(z|\boldsymbol{w}_c)$ according to Eq. (8)\\
        Compute $\mu(\boldsymbol{w}_c)$ and $\sigma(\boldsymbol{w}_c)$ according to Eq. (16), (17)\\
        Sample text embedding $\boldsymbol{z}_c$ from $p(z|\boldsymbol{w}_c)$\\
        Compute uncertainty $\log\sigma^2$ according to Eq. (18)\\
        Compute $\mathcal{L}_{pixel}$ according to Eq. (10)\\
        Compute $\mathcal{L}_{prob}$ according to Eq. (11)\\
        Compute $\mathcal{L}_{KL}$ according to Eq. (12)\\
        Compute total loss $\mathcal{L}$ according to Eq. (13)\\
        Update $\mathbf{P}$ and $\mathcal{M}$ by gradient descent\\}
\end{algorithm}
\vspace{-13pt}

\section{Additional Visualization}
\begin{figure*}[t]
\centering
\begin{subfigure}[t]{0.135\linewidth}
    \makebox[0pt][r]{\makebox[15pt]{\raisebox{13pt}{\rotatebox[origin=c]{90}{\textit{tree}}}}}%
    \includegraphics[width=\textwidth]{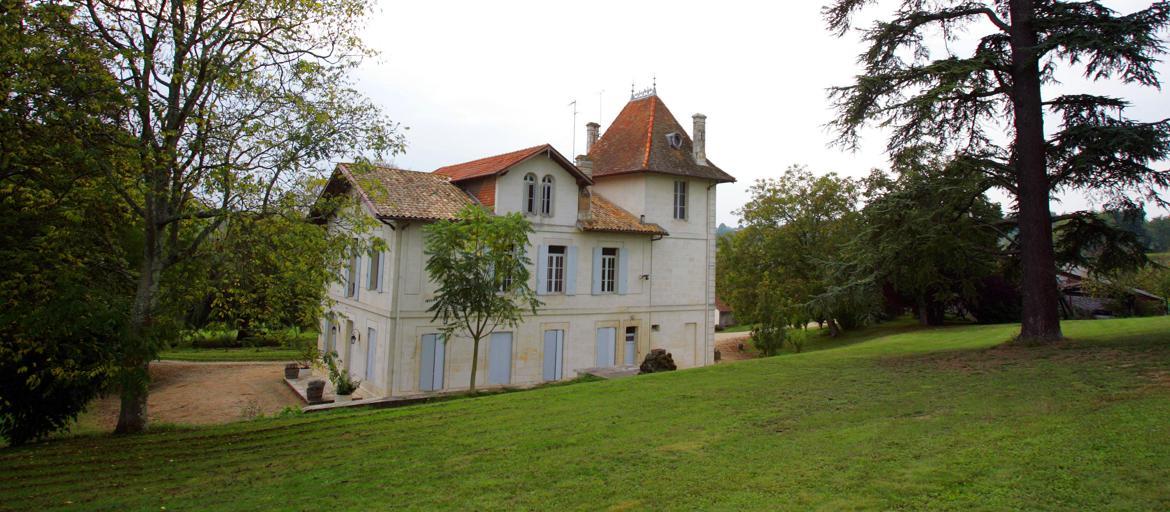}
    \makebox[0pt][r]{\makebox[15pt]{\raisebox{20pt}{\rotatebox[origin=c]{90}{\textit{cabinet}}}}}%
    \includegraphics[width=\textwidth]{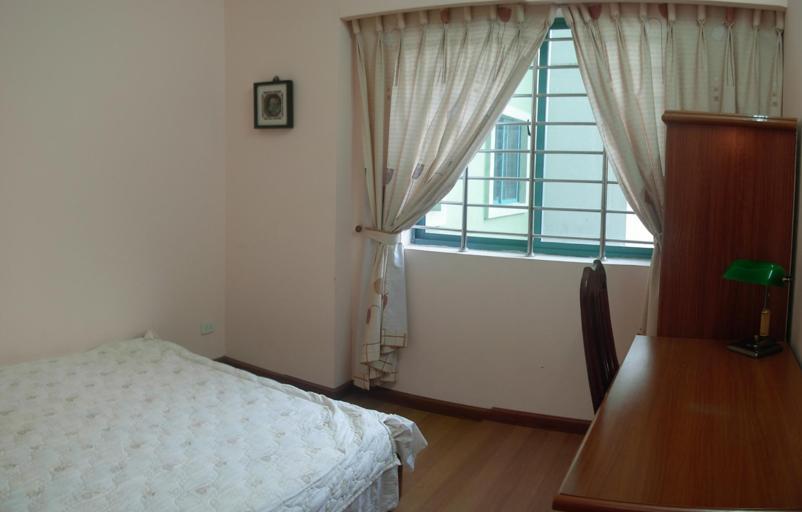}
    \makebox[0pt][r]{\makebox[15pt]{\raisebox{20pt}{\rotatebox[origin=c]{90}{\textit{grass}}}}}%
    \includegraphics[width=\textwidth]{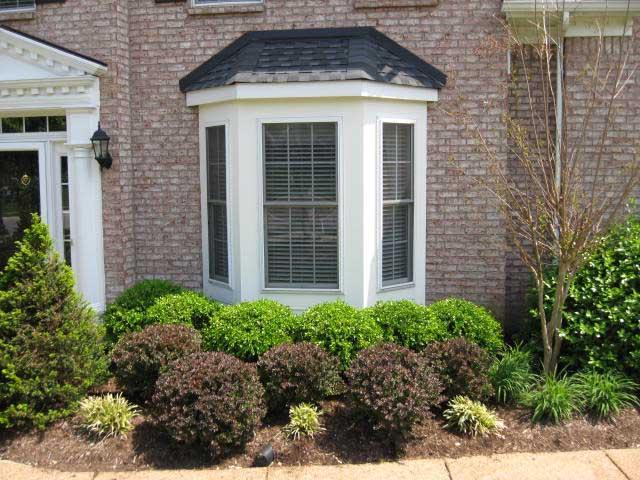}
    \makebox[0pt][r]{\makebox[15pt]{\raisebox{20pt}{\rotatebox[origin=c]{90}{\textit{bear}}}}}%
    \includegraphics[width=\textwidth]{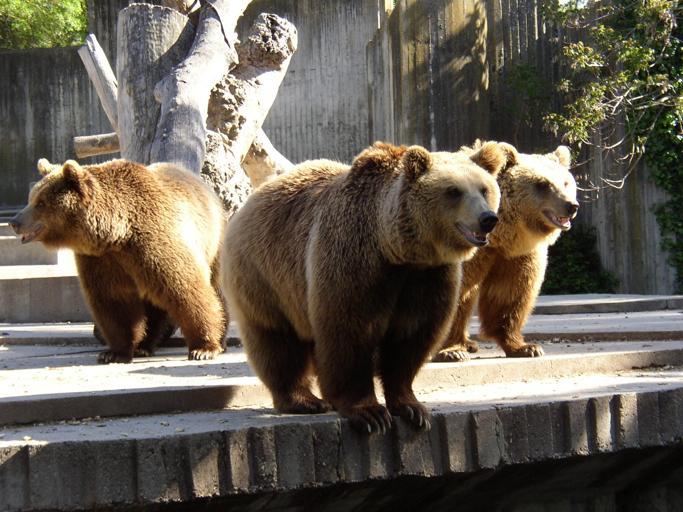}
    \makebox[0pt][r]{\makebox[15pt]{\raisebox{25pt}{\rotatebox[origin=c]{90}{\textit{person}}}}}%
    \includegraphics[width=\textwidth]{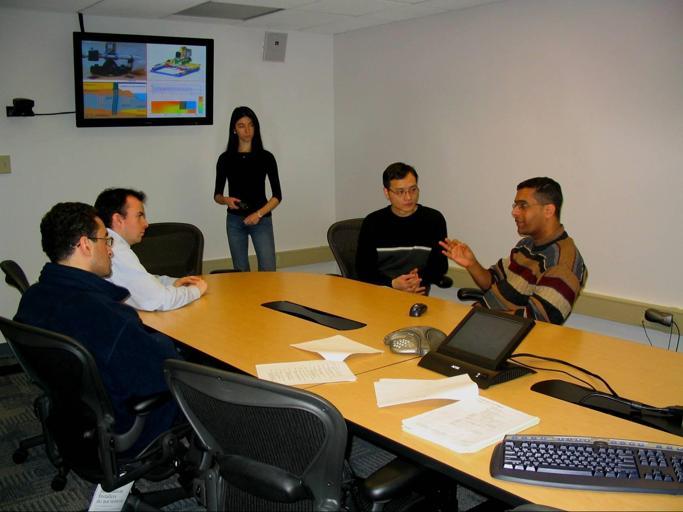}
    \makebox[0pt][r]{\makebox[15pt]{\raisebox{25pt}{\rotatebox[origin=c]{90}{\textit{lamp}}}}}%
    \includegraphics[width=\textwidth]{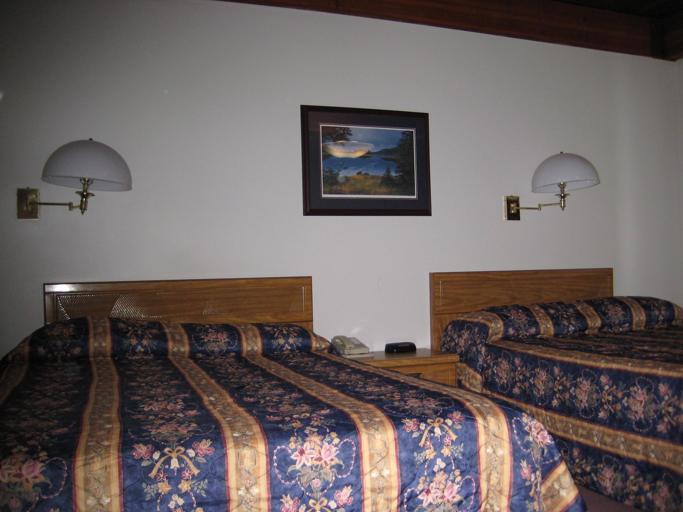}
    \makebox[0pt][r]{\makebox[15pt]{\raisebox{20pt}{\rotatebox[origin=c]{90}{\textit{sofa}}}}}%
    \includegraphics[width=\textwidth]{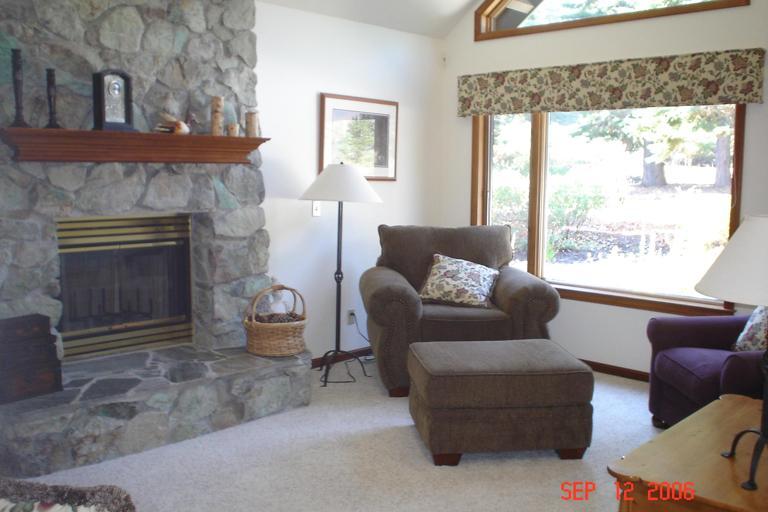}

    \caption{Image}
\end{subfigure}
\begin{subfigure}[t]{0.135\linewidth}
    \includegraphics[width=\textwidth]{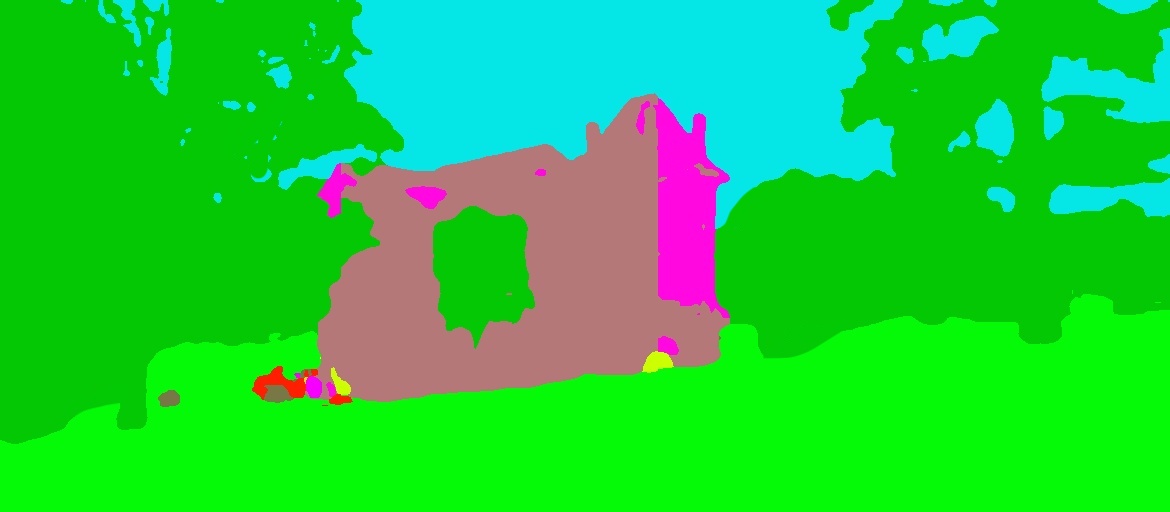}
    \includegraphics[width=\textwidth]{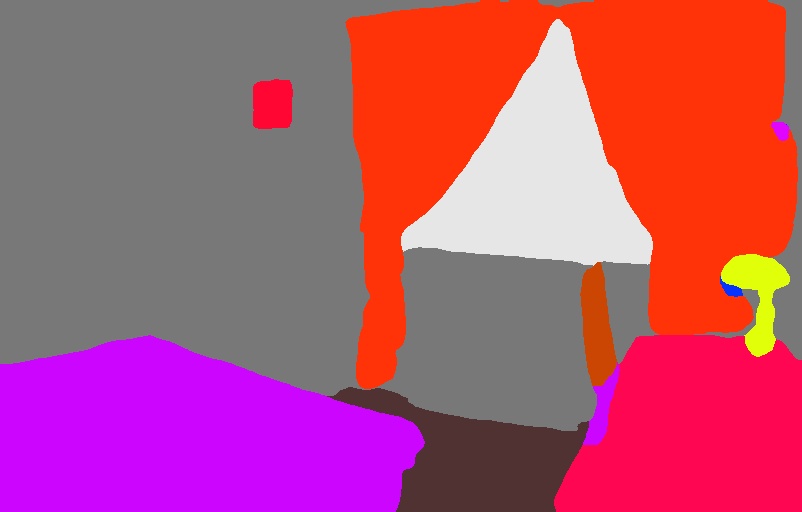}
    \includegraphics[width=\textwidth]{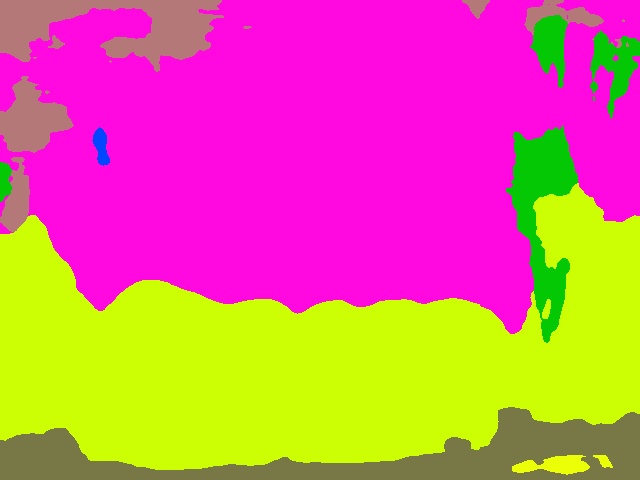}
    \includegraphics[width=\textwidth]{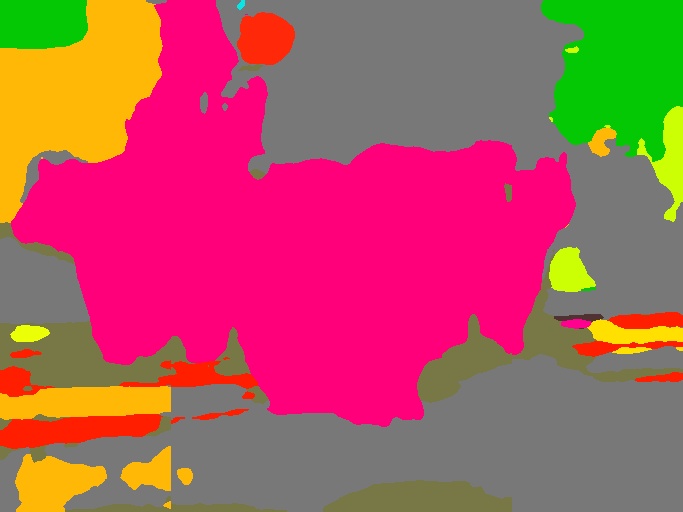}
    \includegraphics[width=\textwidth]{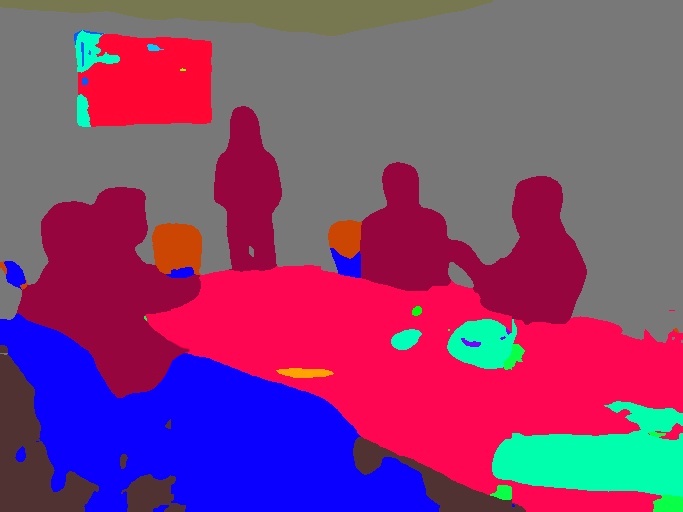}
    \includegraphics[width=\textwidth]{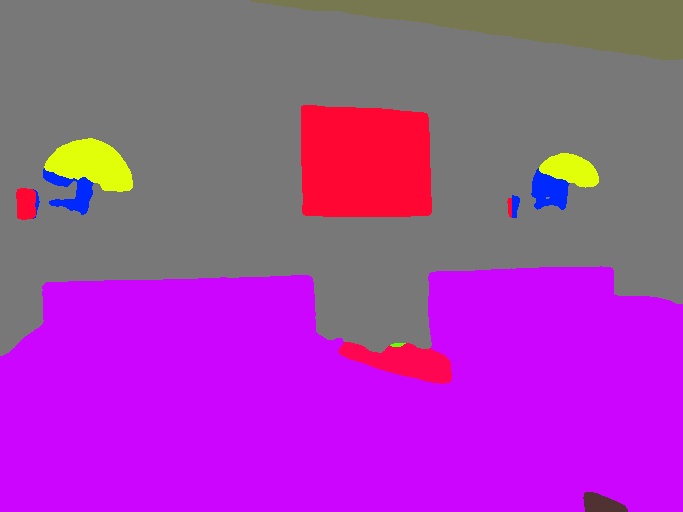}
    \includegraphics[width=\textwidth]{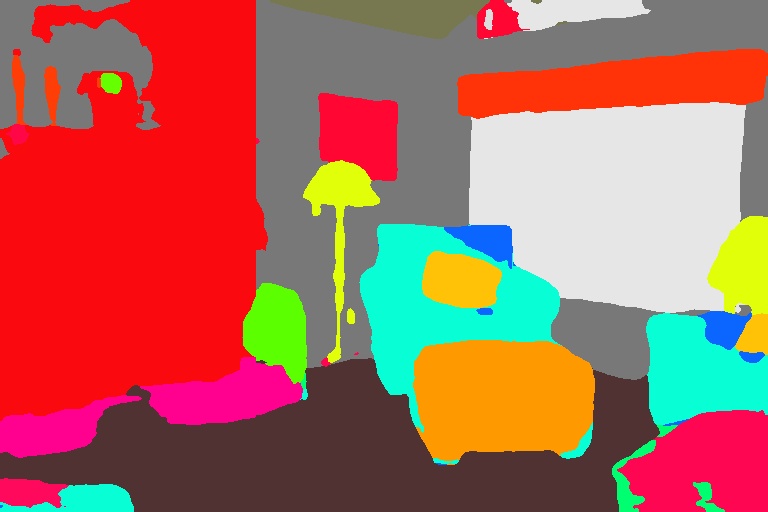}
    \caption{DenseCLIP~\cite{rao2022denseclip}}
\end{subfigure}
\begin{subfigure}[t]{0.135\linewidth}
    \includegraphics[width=\textwidth]{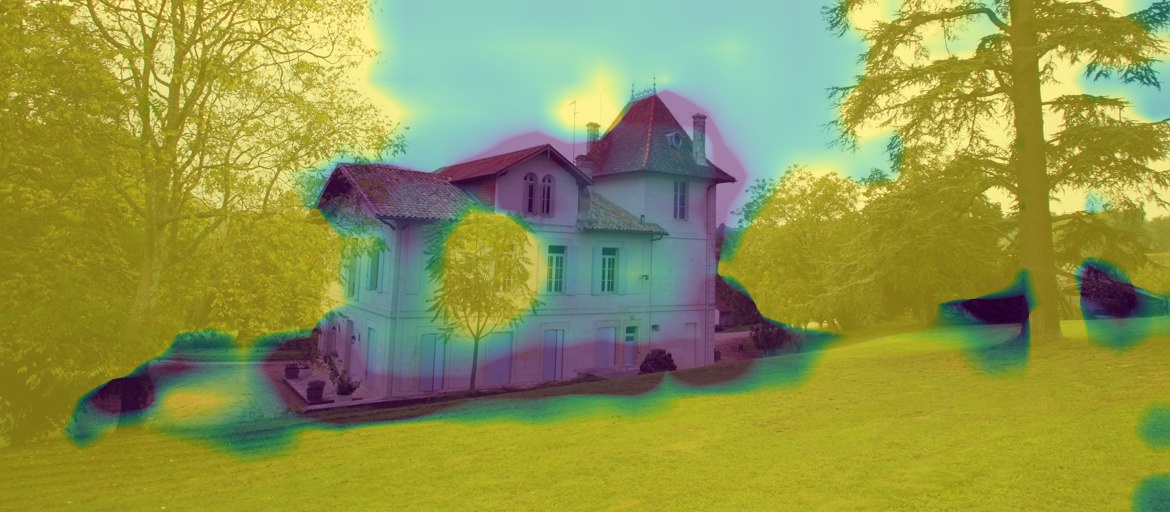}
    \includegraphics[width=\textwidth]{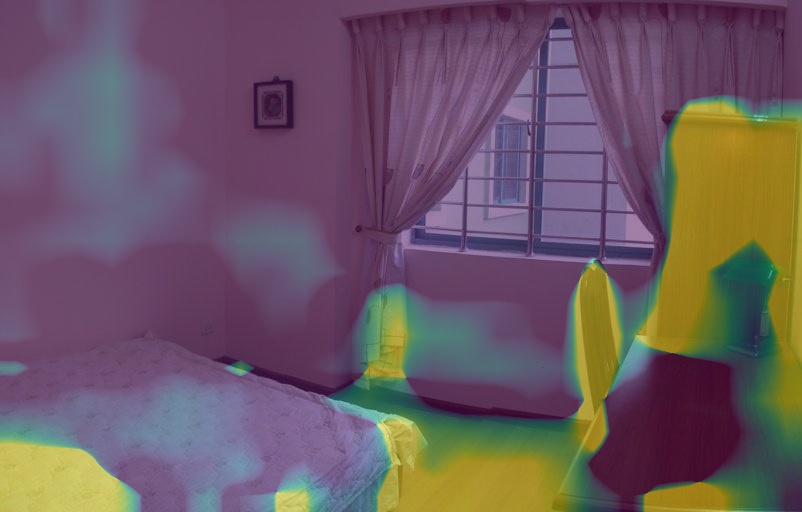}
    \includegraphics[width=\textwidth]{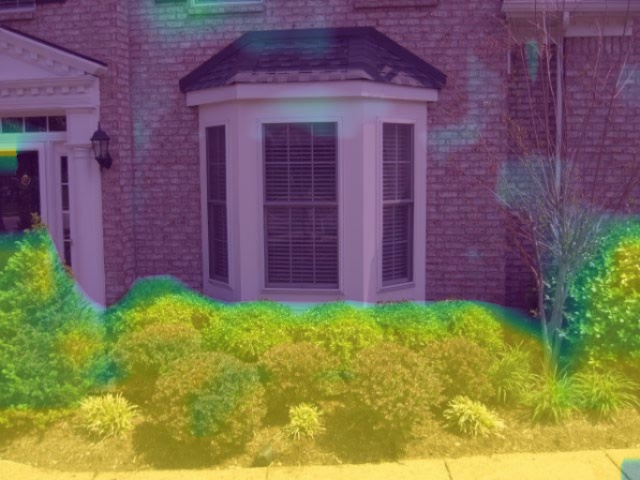}
    \includegraphics[width=\textwidth]{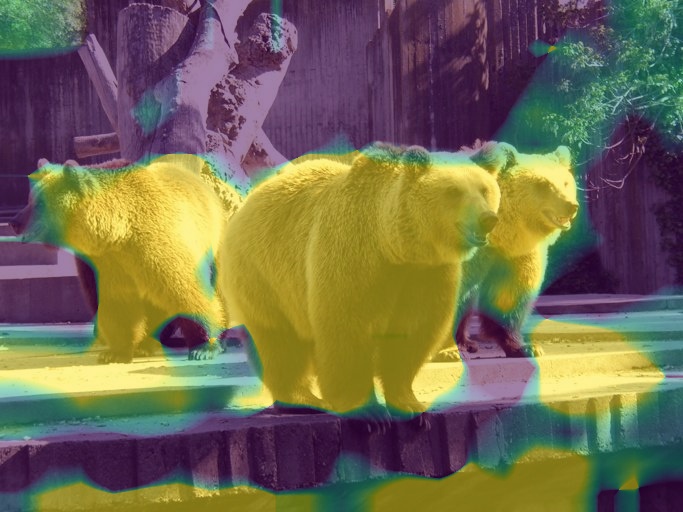}
    \includegraphics[width=\textwidth]{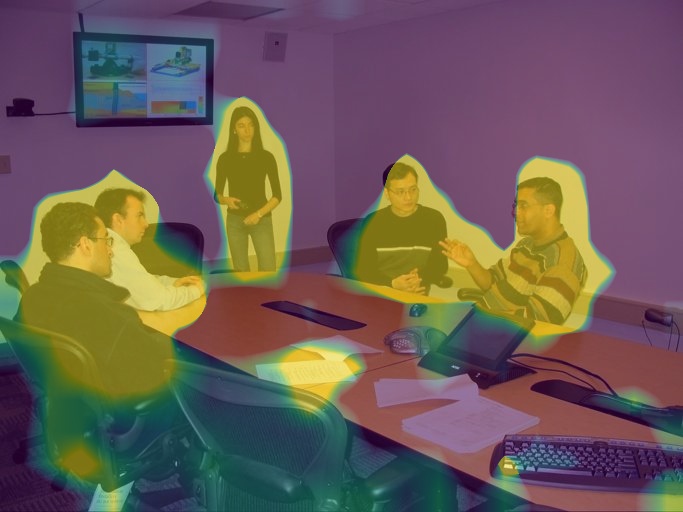}
    \includegraphics[width=\textwidth]{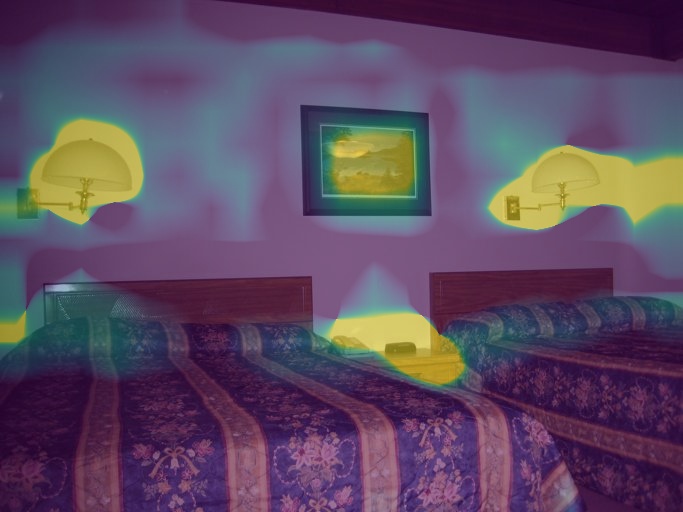}
    \includegraphics[width=\textwidth]{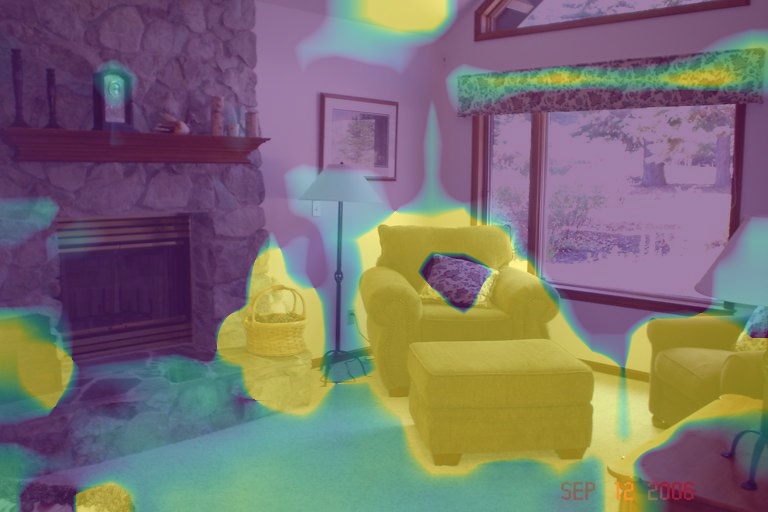}
    \caption{\texttt{Samples 1}}
\end{subfigure}
\begin{subfigure}[t]{0.135\linewidth}
    \includegraphics[width=\textwidth]{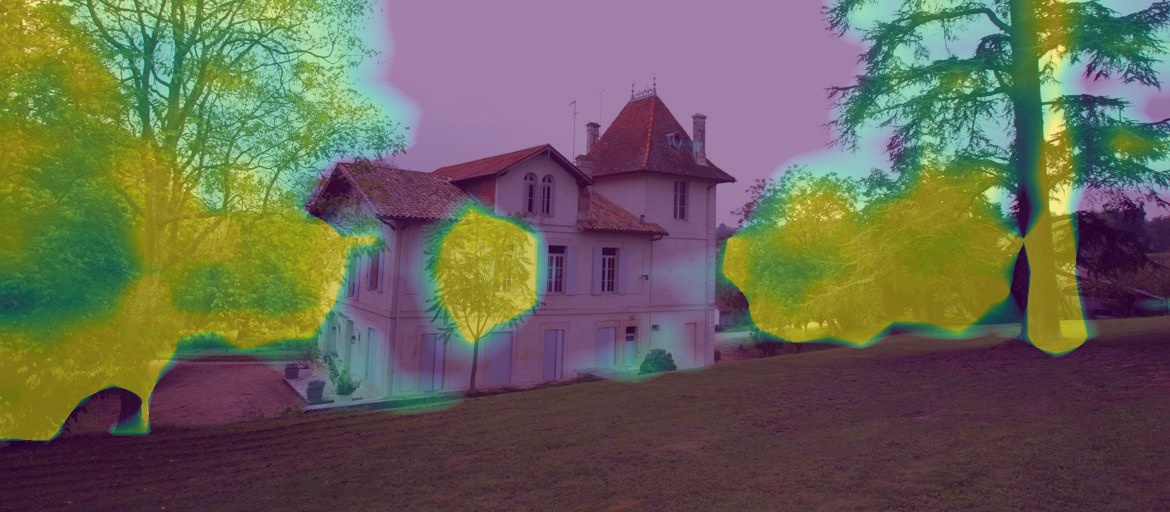}
    \includegraphics[width=\textwidth]{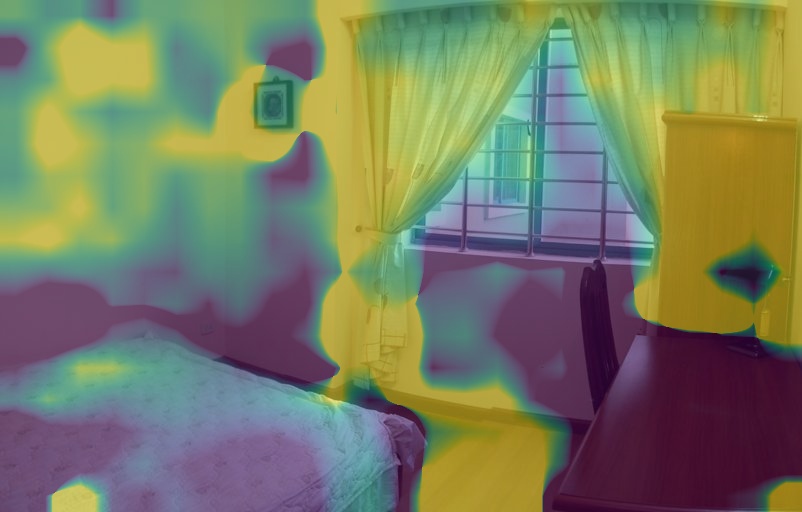}
    \includegraphics[width=\textwidth]{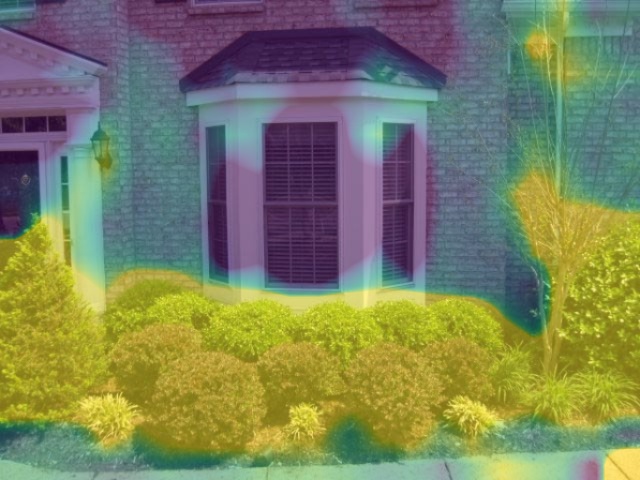}
    \includegraphics[width=\textwidth]{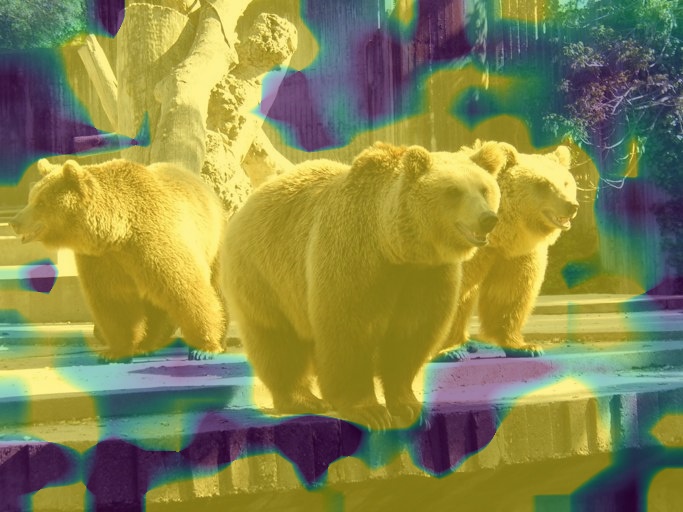}
    \includegraphics[width=\textwidth]{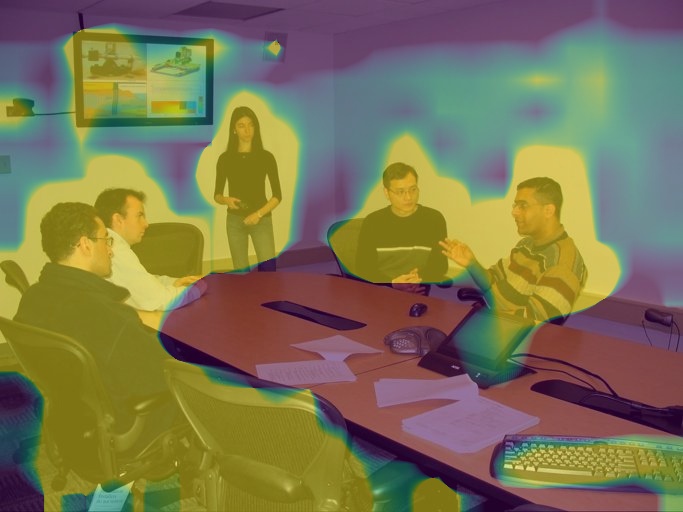}
    \includegraphics[width=\textwidth]{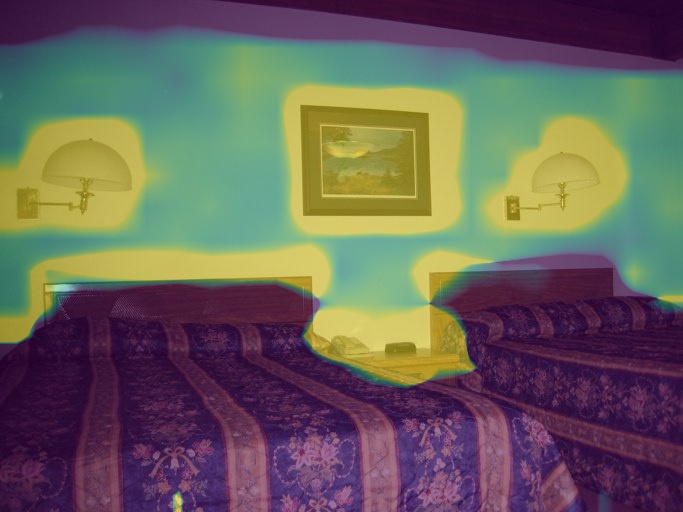}
    \includegraphics[width=\textwidth]{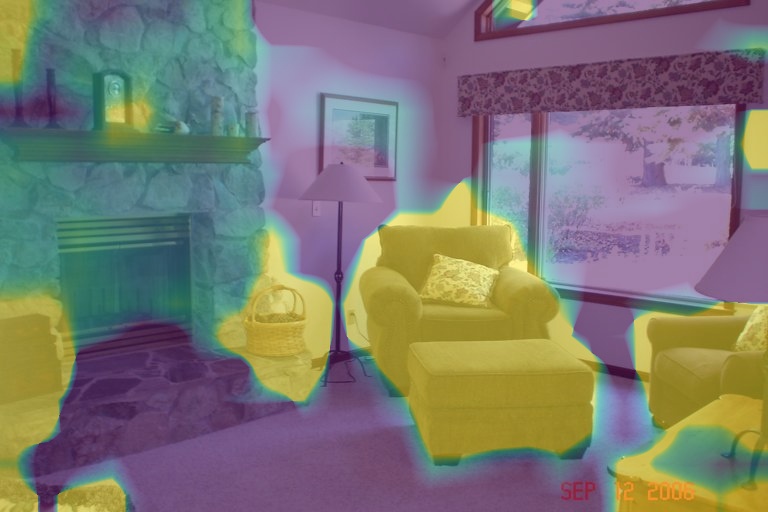}
    \caption{\texttt{Samples 2}}
\end{subfigure}
\begin{subfigure}[t]{0.135\linewidth}
    \includegraphics[width=\textwidth]{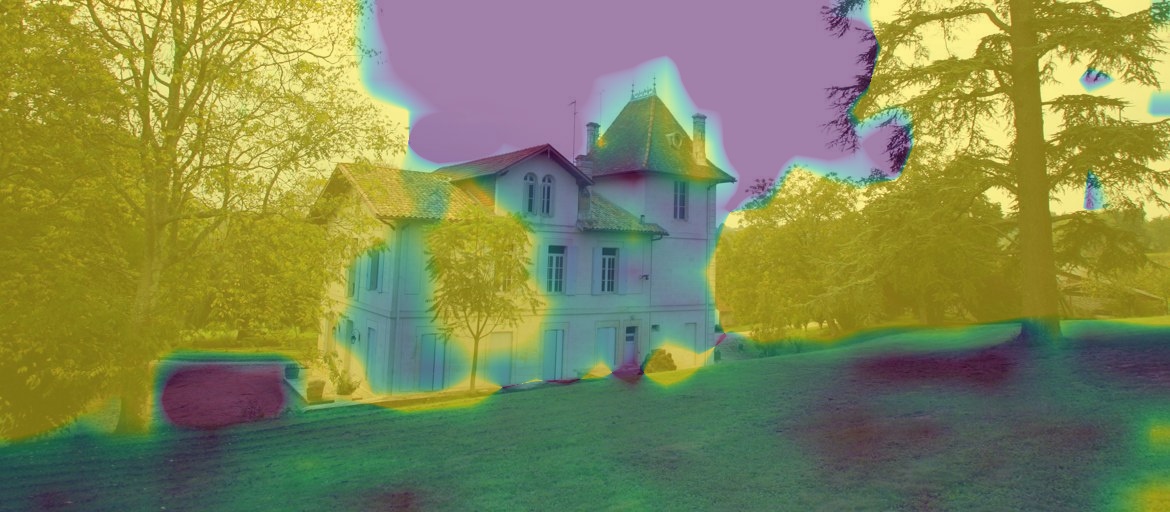}
    \includegraphics[width=\textwidth]{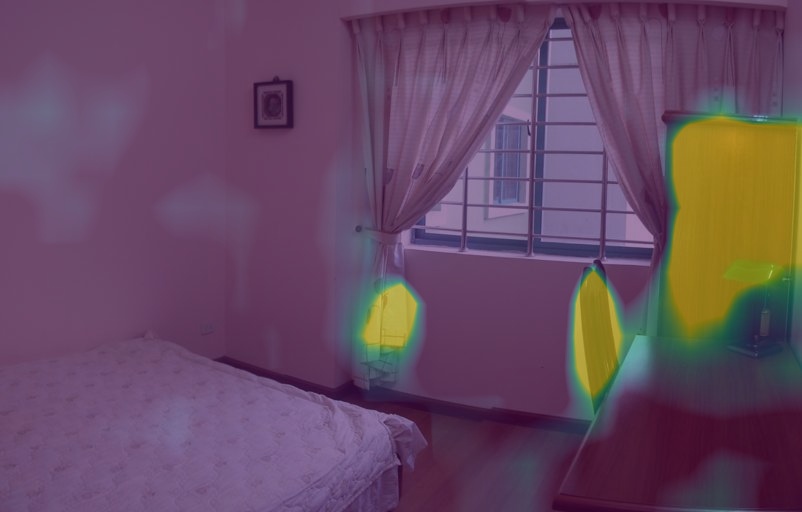}
    \includegraphics[width=\textwidth]{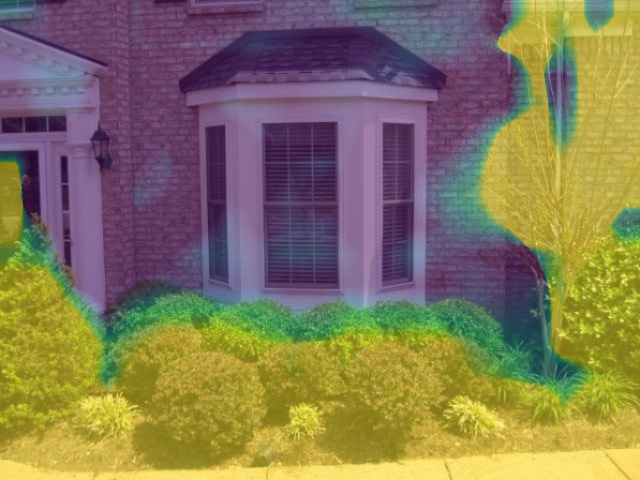}
    \includegraphics[width=\textwidth]{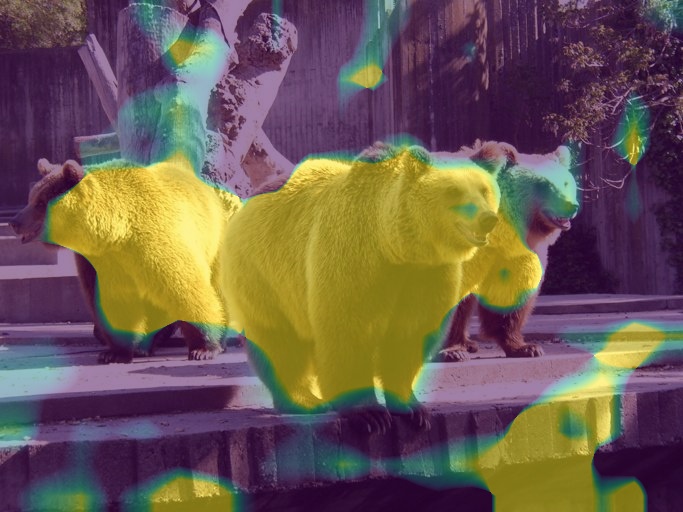}
    \includegraphics[width=\textwidth]{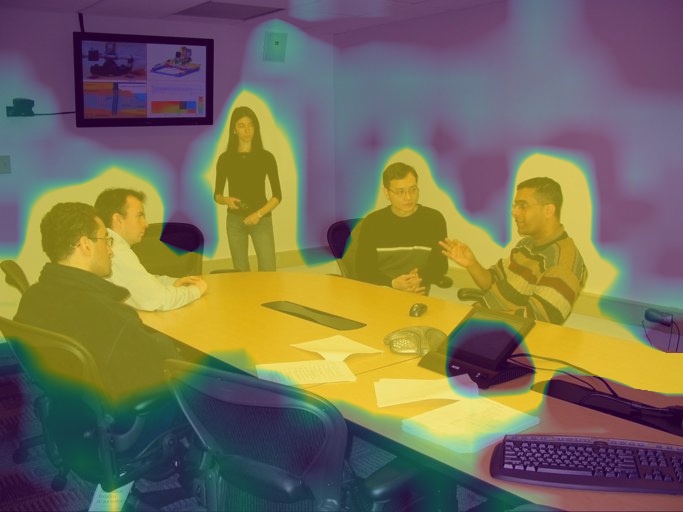}
    \includegraphics[width=\textwidth]{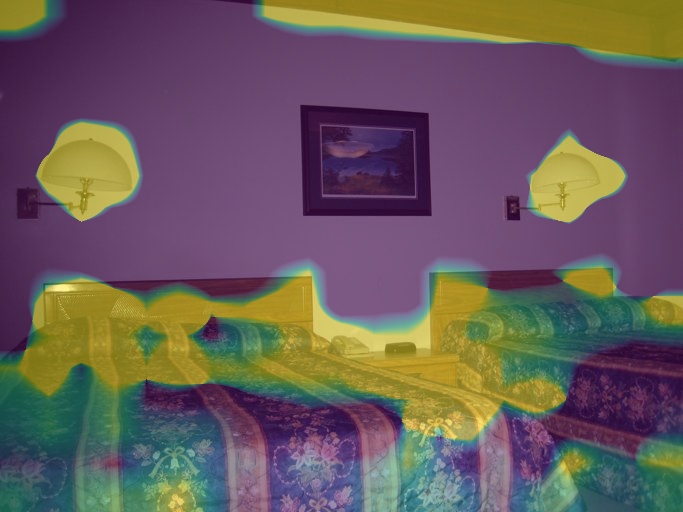}
    \includegraphics[width=\textwidth]{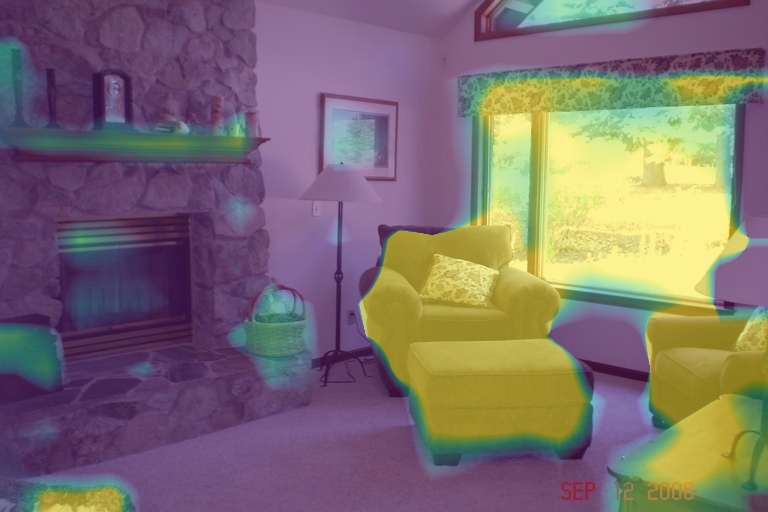}
    \caption{\texttt{Samples 3}}
\end{subfigure}
\begin{subfigure}[t]{0.135\linewidth}
    \includegraphics[width=\textwidth]{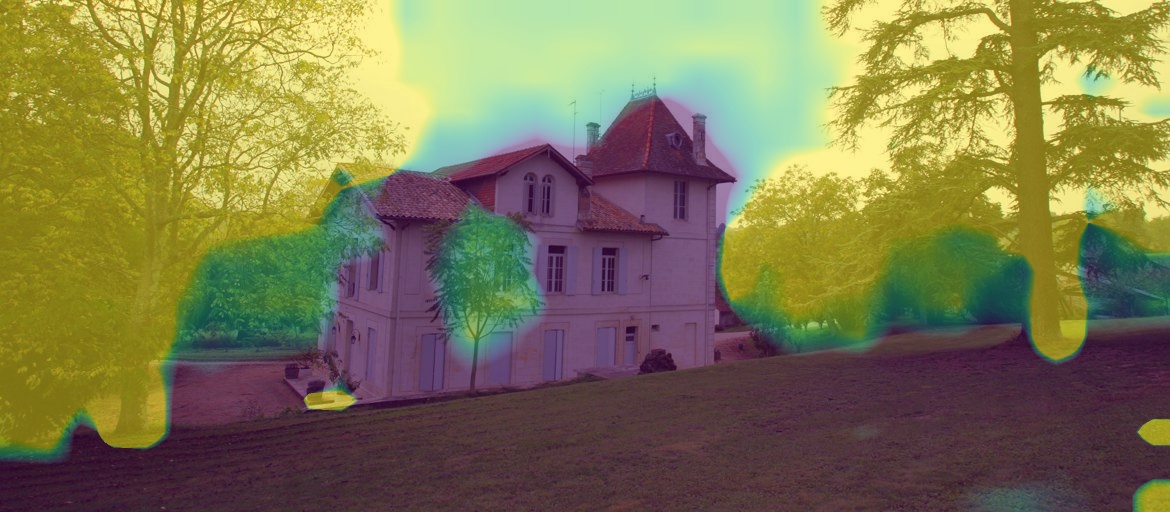}
    \includegraphics[width=\textwidth]{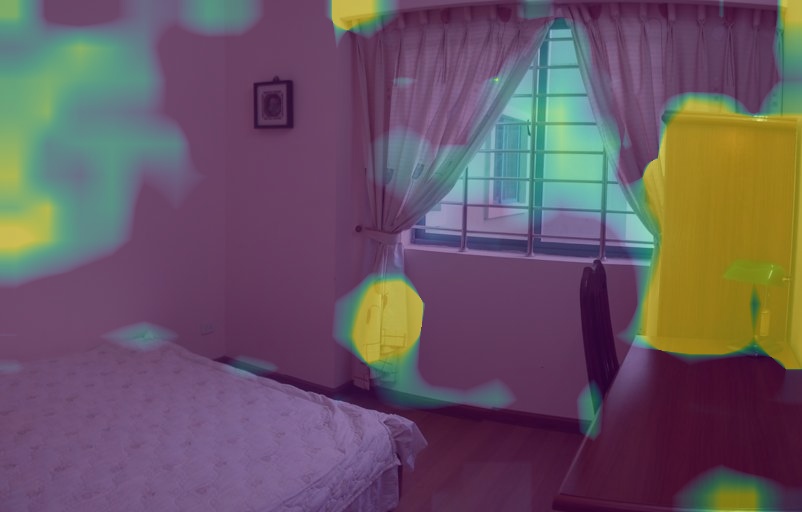}
    \includegraphics[width=\textwidth]{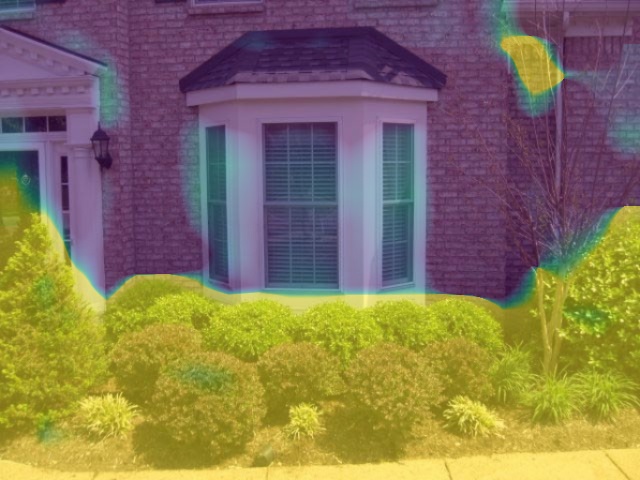}
    \includegraphics[width=\textwidth]{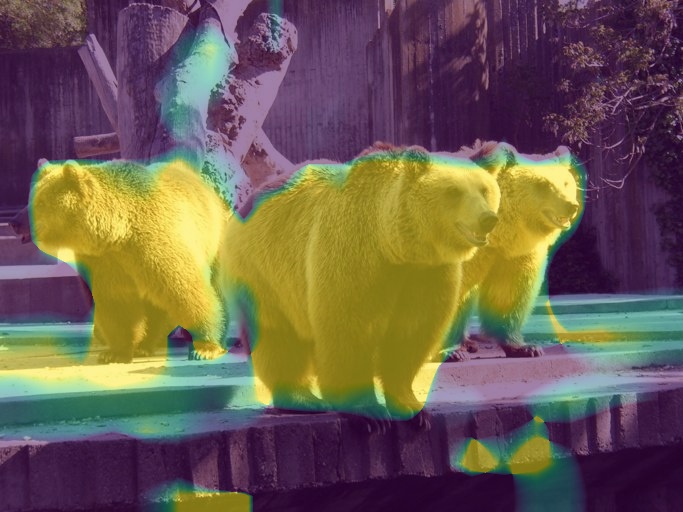}
    \includegraphics[width=\textwidth]{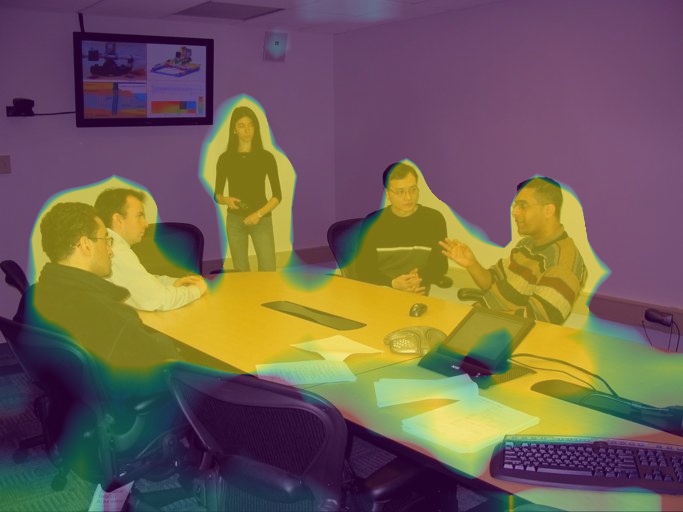}
    \includegraphics[width=\textwidth]{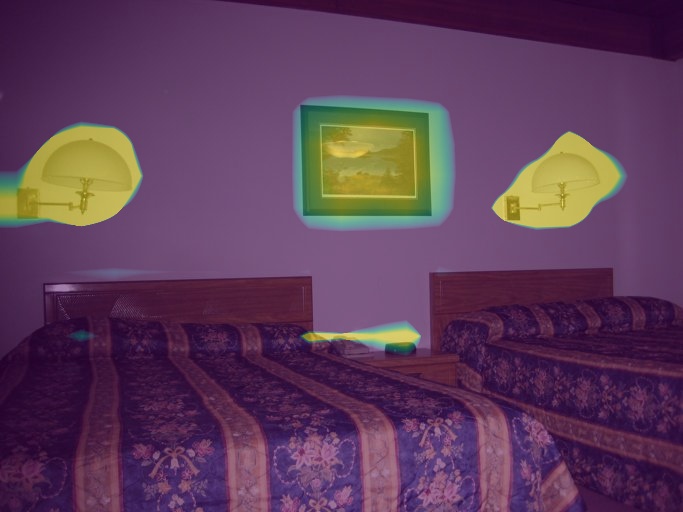}
    \includegraphics[width=\textwidth]{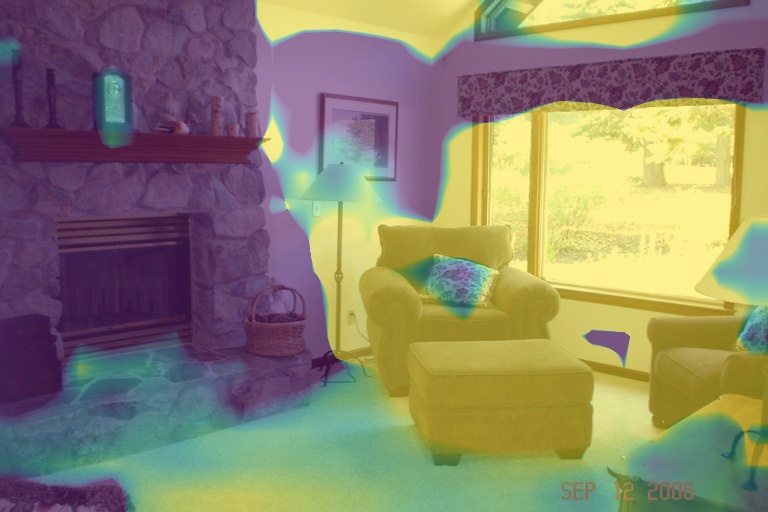}
    \caption{\texttt{Samples 4}}
\end{subfigure}
\begin{subfigure}[t]{0.135\linewidth}
    \includegraphics[width=\textwidth]{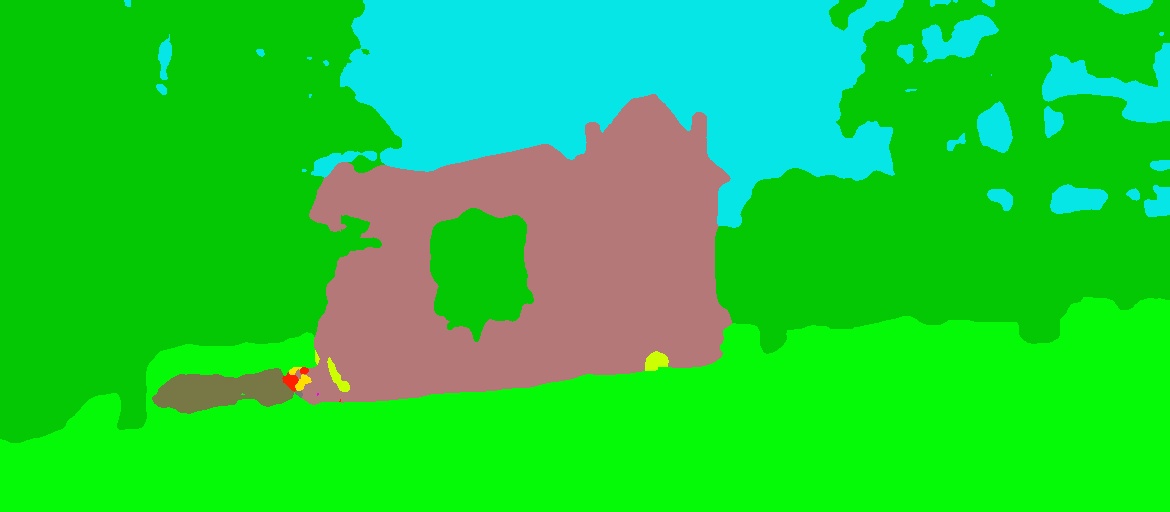}
    \includegraphics[width=\textwidth]{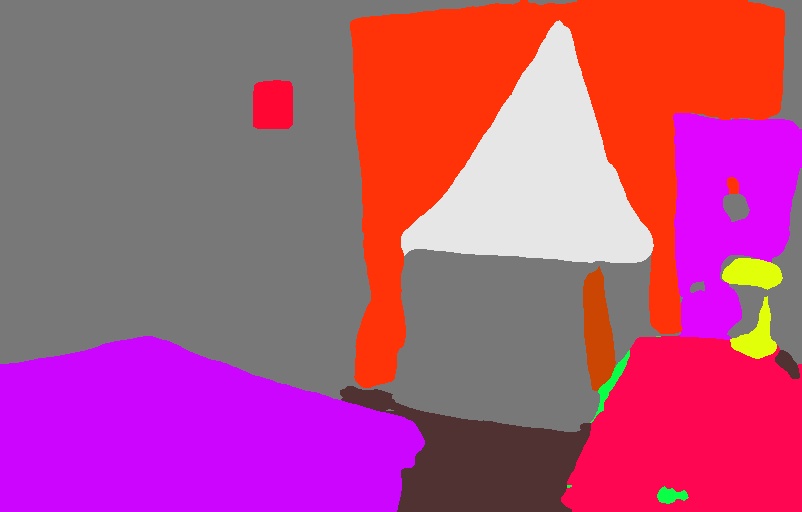}
    \includegraphics[width=\textwidth]{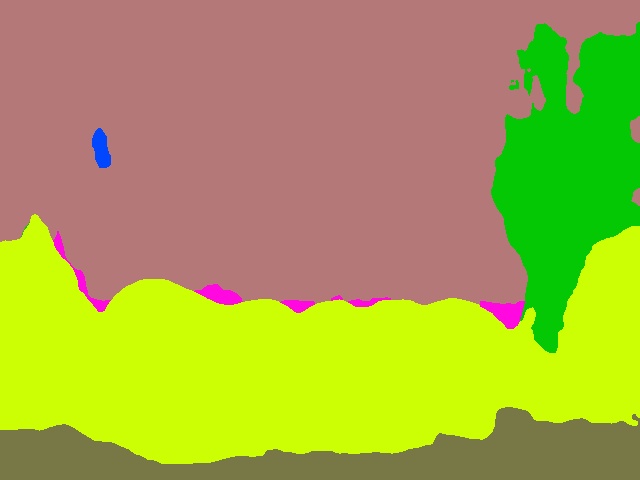}
    \includegraphics[width=\textwidth]{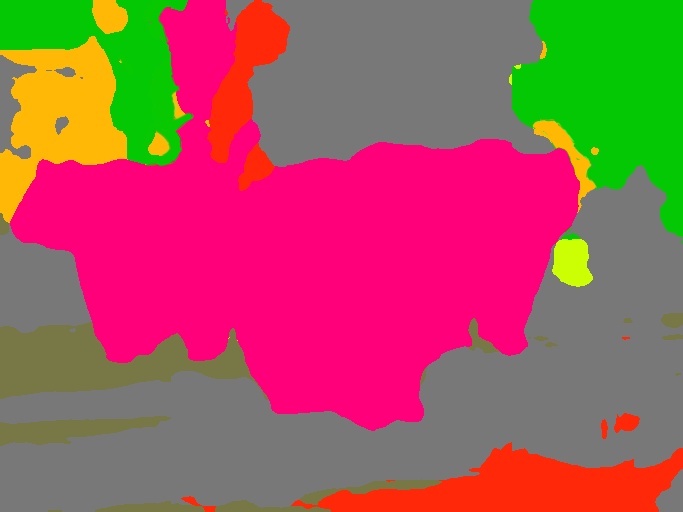}
    \includegraphics[width=\textwidth]{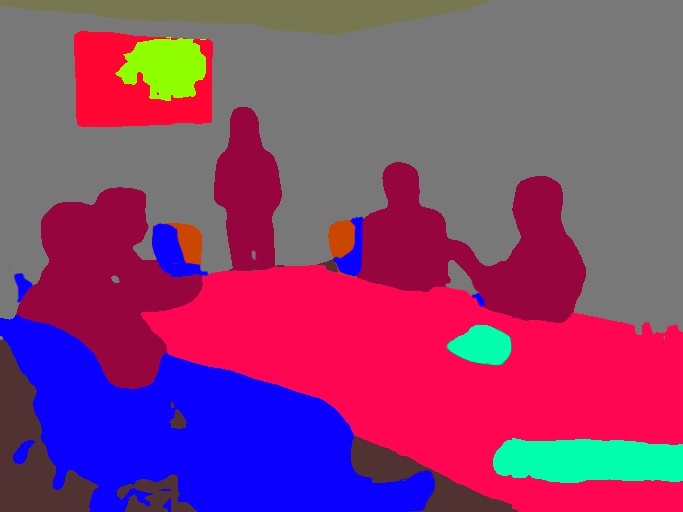}
    \includegraphics[width=\textwidth]{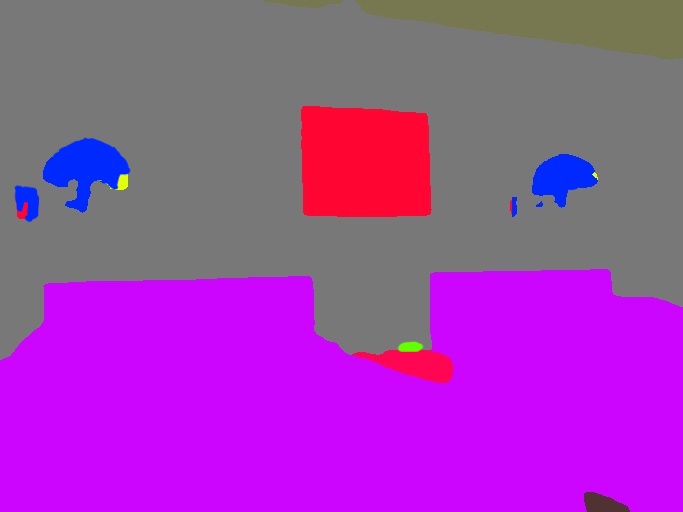}
    \includegraphics[width=\textwidth]{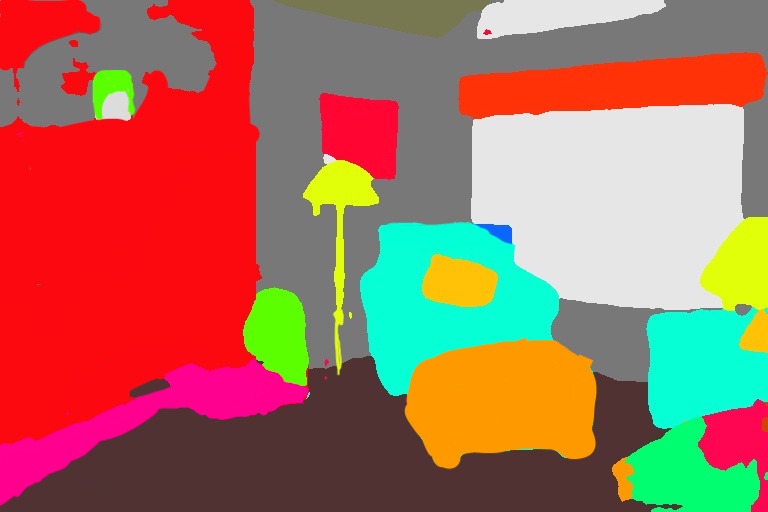}
    \caption{PPL(ours)}
\end{subfigure}
\vspace{-4pt}
\caption{\textbf{\textbf{Visualization of activation maps and segmentation results}}. We visualize the activation maps (c), (d) (e), and (f) of sampled representation of different classes indicated on the left side, with $K=3$ on the ADE20k dataset~\cite{zhou2019semantic}. We report qualitative results of segmentation of both (b) DenseCLIP~\cite{rao2022denseclip}, (g) PPL(ours).}
\label{suppl:act}
\vspace{-4pt}
\end{figure*}

In this section, we provide more visualization results of our method and comparison our method with DenseCLIP~\cite{rao2022denseclip}.
As shown if Fig.~\ref{suppl:act}, we showed that each similarity map with different visual context represent the target class object as different ways.
Specifically, combining different similarity maps remove undesirable prediction and improves performance.
We report the qualitative results with given score maps compared to DenseCLIP~\cite{rao2022denseclip}.
\label{sec:add1}
\newpage
\clearpage

\end{document}